\pdfoutput=1

\documentclass[11pt]{article}

\usepackage[]{acl}

\usepackage{times}
\usepackage{latexsym}
\usepackage{nicematrix}
\usepackage{amsmath}
\usepackage{amssymb}
\usepackage{algpseudocode}
\usepackage{algorithm}
\usepackage[T1]{fontenc}

\usepackage[utf8]{inputenc}
\usepackage{svg}
\usepackage{subcaption}
\usepackage{xspace}
\usepackage{tabularx}
\usepackage{euflag}

\usepackage{microtype}
\usepackage{booktabs}

\usepackage{xcolor, colortbl}
\definecolor{nice-red}{HTML}{E41A1C}
\definecolor{nice-orange}{HTML}{FF7F00}
\definecolor{nice-yellow}{HTML}{FFC020}
\definecolor{nice-green}{HTML}{4DAF4A}
\definecolor{nice-blue}{HTML}{377EB8}
\definecolor{nice-purple}{HTML}{984EA3}
\definecolor{gray-blue}{HTML}{d1e5f0}
\definecolor{gray-red}{HTML}{edb7bd}

\usepackage{todonotes}
\makeatletter
\newcommand*\iftodonotes{\if@todonotes@disabled\expandafter\@secondoftwo\else\expandafter\@firstoftwo\fi}
\makeatother

\newcommand{\gbc}{\cellcolor{gray-blue}}

\newcommand{\rparagraph}[1]{\vspace{1.4mm}\noindent\textbf{#1.}}
\newcommand{\sparagraph}[1]{\vspace{0.0mm}\noindent\textbf{#1.}}

\title{
FUN with Fisher: Improving Generalization of Adapter-Based Cross-lingual Transfer with Scheduled Unfreezing
}

\author{\bf Chen Cecilia Liu$^1$, Jonas Pfeiffer$^{2}$,
{\bf Ivan Vuli\'{c}$^{3}$, Iryna Gurevych$^{1}$ } \\
$^1$Ubiquitous Knowledge Processing Lab,\\ Department of Computer Science and Hessian Center for AI (hessian.AI), \\ Technical University of Darmstadt \\
  $^2$Google DeepMind \\
$^3$Language Technology Lab, University of Cambridge \hspace{0.5em} \\
{\url{www.ukp.tu-darmstadt.de}} \\
}

\DeclareMathOperator{\E}{\mathbb{E}}
\DeclareMathOperator{\Tr}{tr}

\begin{document}
\maketitle
\begin{abstract}

Standard fine-tuning of language models typically performs well on \textit{in-distribution data}, but suffers with generalization to \textit{distribution shifts}. In this work, we aim to improve the generalization of adapter-based cross-lingual task transfer where such cross-language distribution shifts are imminent. We investigate scheduled unfreezing algorithms---originally proposed to mitigate catastrophic forgetting in transfer learning---for fine-tuning task adapters. Our experiments show that scheduled unfreezing methods close the gap to full fine-tuning and achieve stronger cross-lingual transfer performance, suggesting that these methods can go beyond just mitigating catastrophic forgetting. Next, aiming to understand these empirical findings, we investigate the learning dynamics of scheduled unfreezing using Fisher Information. Our experiments reveal that scheduled unfreezing induces different learning dynamics compared to standard fine-tuning, and provide evidence that the dynamics of Fisher Information during training correlate with cross-lingual generalization performance. We additionally propose a general scheduled unfreezing algorithm that achieves an average of 2 points improvement over four datasets compared to standard fine-tuning and provides empirical evidence for a theory-based justification of the heuristic unfreezing schedule for adapter training.~\footnote{ \url{https://github.com/UKPLab/naacl2024-fun}}
\end{abstract}

\section{Introduction}

In the standard cross-lingual task transfer setup, a typical and often valid assumption is that only English data is available for fine-tuning and validation of a pretrained multilingual model, due to resource constraints in many languages~\cite{xtreme}. However, models trained in this setup also need to generalize well to text inputs provided in other languages: a requirement that can be seen as an extreme but natural case of \emph{distribution shifts} generalization~\cite{ramponi-plank-2020-neural}.  

Parameter-efficient fine-tuning methods such as adapters~\cite{pmlr-v97-houlsby19a, pmlr-v97-stickland19a, bapna-firat-2019-simple} with separate language and task components are often used to achieve effective cross-lingual transfer, especially to low-resource languages~\cite{mad-x,pfeiffer-etal-2021-unks, ansell-etal-2021-mad-g, parovic-etal-2022-bad}. These adapters insert a small number of trainable parameters into a frozen pretrained multilingual language model (e.g., mBERT, \citealp{devlin-etal-2019-bert}; XLM-R, \citealp{conneau-etal-2020-unsupervised}) to achieve positive transfer while avoiding catastrophic forgetting (CF, ~\citealp{MCCLOSKEY1989109}) of previously learnt knowledge after adapting to new tasks.\footnote{Adapters bypass this issue since the pretrained model is kept frozen, with its original weights unchanged.} In other words, adapters enable \emph{catastrophic forgetting free} learning, referred to as \emph{CF-free} in this paper. While efficient, adapter methods often incur a cross-lingual performance gap when compared to full model fine-tuning.

\textit{Gradual unfreezing} (GU) is a technique which unfreezes layers of deep neural network models from top to bottom during training~\cite{gu}. GU was previously proposed for general transfer learning of in-distribution (ID) data in monolingual contexts in NLP, and has been predominantly applied to full fine-tuning. More recently, `Linear-Probing-then-Fine-Tuning' (LPFT, \citealp{kumar2022finetuning}) was proposed for transfer learning of both ID and distribution-shifted evaluation data using full fine-tuning in computer vision. LPFT first trains the classification layer only, and then the full model. The main notion connecting these methods is training different layers of a neural network by unfreezing layers at different times (i.e., with a \textit{schedule}). Designed to mitigate CF, these \textit{`scheduled unfreezing'} methods have shown promising transfer learning results. However, it is unclear whether scheduled unfreezing can do more than just mitigate CF, and benefit CF-free methods and cross-lingual transfer (which is a different type of distribution shift than previously studied).

In this work, we begin by asking the following question: Do scheduled unfreezing methods improve cross-lingual transfer and close the gap to full fine-tuning in the CF-free setting? We use scheduled unfreezing to train task adapters following the standard adapter-based cross-lingual transfer setup of~\citet{mad-x}. We find that scheduled unfreezing enhances generalization, bridging the gap to full fine-tuning, confirming our hypothesis that, since cross-lingual transfer can be seen as a form of distribution shifts, methods such as GU are effective, even in the CF-free setting. Our results suggest that there indeed is more to the original scheduled unfreezing training than just mitigating catastrophic forgetting~\cite{gu}.

We further analyze the learning dynamics during training, with a particular focus on GU, using the trace of the Fisher Information Matrix (\citealp{achille2022critical}, denoted by $\Tr(F)$ henceforth). Our experiments reveal \textbf{1)} that scheduled unfreezing changes the dynamics of $\Tr(F)$ during training, \textbf{2)} that $\Tr(F)$ is a potential proxy for studying cross-lingual generalization.

Based on our analysis, we then propose an automatic scheduled unfreezing algorithm based on maximizing the $\Tr(F)$ (termed \textbf{F}isher \textbf{Un}freezing or FUN), to generalize from previous heuristic-based methods. FUN achieves comparable results to heuristic-based methods and provides empirical evidence that GU may implicitly maximize $\Tr(F)$ during training in our experimental setting. 

\rparagraph{Contributions} In sum, our contributions are as follows. \textbf{1)} To the best of our knowledge, we are the first to demonstrate that scheduled unfreezing in adapter training for cross-lingual transfer closes the performance gap to full model fine-tuning. \textbf{2)} We present a generalized scheduled unfreezing framework that encompasses several existing methods, allowing easy extensions to new algorithms. \textbf{3)} We demonstrate that Fisher Information is an effective tool for studying generalization in cross-lingual transfer and find that the dynamics of Fisher Information correlate with cross-lingual transfer results. \textbf{4)} We propose a $\Tr(F)$-based scheduled unfreezing method (FUN), which achieves comparable performance to heuristic methods and indicates that GU achieves good generalization in adapter training.

\begin{figure*}[!t]
    \centering
    \includegraphics[width=0.9\textwidth]{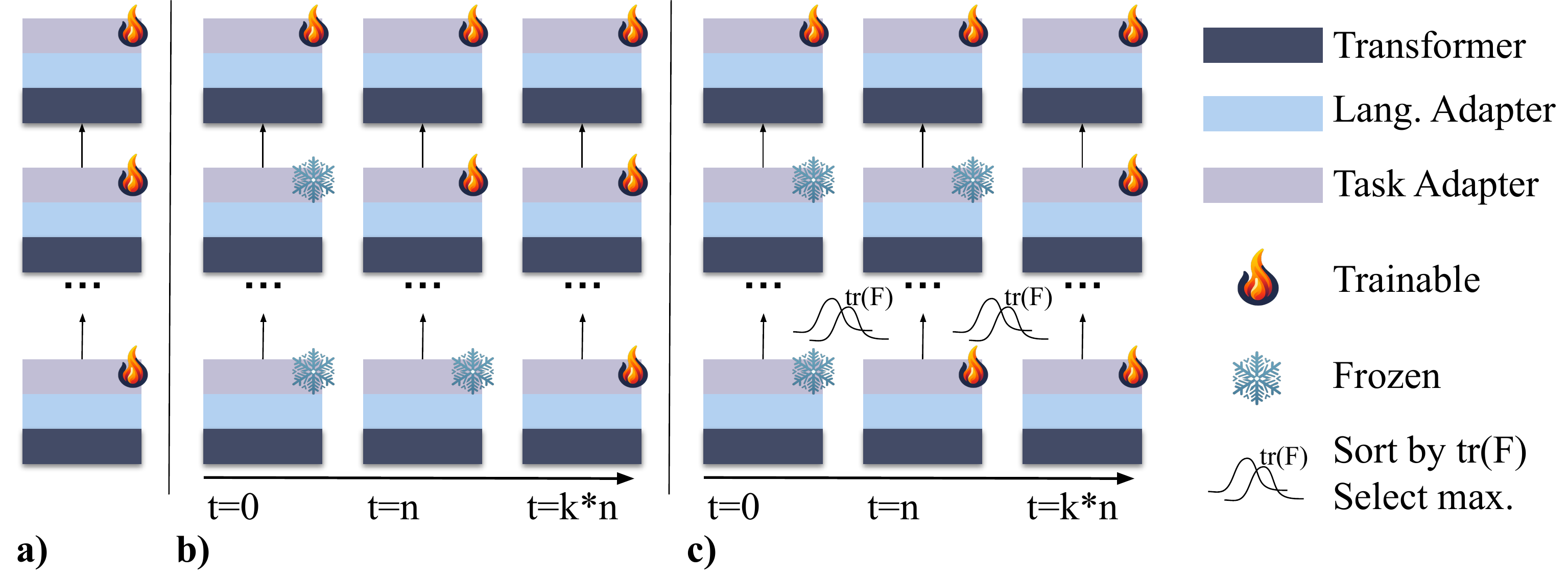}
    \caption{\textbf{a)} Standard, \textbf{b)} Gradual unfreezing versus \textbf{c)} $\Tr(F)$-based scheduled unfreezing for training task adapters in adapter-based cross-lingual transfer. The classifier is not shown and is always trainable. All other components excluding task adapters, such as the original parameters of the base model and language adapters, are always frozen.}
    \label{fig:my_label}
\end{figure*}

\section{Related Work}

\sparagraph{Scheduled Unfreezing}
\citet{gu} propose Gradual Unfreezing (GU) to mitigate catastrophic forgetting when transferring a pretrained model for monolingual downstream tasks in non-Transformer architectures. \citet{raffel2022t5} study GU for transferring full fine-tuning Transformers to in-distribution tasks, and empirically conclude that GU may not be effective. Recently, \citet{kumar2022finetuning} (LPFT, Linear Probing then Fine-Tuning) show promising results for distribution-shifted transfer in the full fine-tuning setting for computer vision tasks. \citet{parameffhead} extend LPFT for full fine-tuning to NLP. \citet{liang2022surgical} show that fine-tuning different parts of a network helps with generalization under different types of distribution shifts in computer vision. Different from our work, the prior work has not studied scheduled unfreezing methods for cross-lingual transfer in a CF-free setting.

\rparagraph{Fisher Information and Generalization} 
Fisher Information~\cite{Fisher1925TheoryOS} is an established concept in optimization theory and practice~\cite{natural-grad}, e.g., to measure parameter importance~\cite{kirkpatrick17} or for pruning~\cite{fishprune}. \citet{dnncritical} study learning dynamics\footnote{Different from training dynamics~\cite{swayamdipta-etal-2020-dataset} which focus on the data, learning dynamics focus on the model and the optimization process.} in neural network training with Fisher Information. \citet{achille2019time} show that Fisher Information correlates with generalization in computer vision and non-Transformer architectures. \citet{jastrzebski2021catastrophic} propose to regularize Fisher Information during the training of a neural network for better in-distribution generalization in computer vision. \citet{childtuning,sung2021training} create sparse masks using Fisher Information for better parameter-efficient tuning. Concurrently, \citet{lodha-etal-2023-surgical} uses Fisher Information selecting layers for fine-tuning transformers. We study Fisher Information in adapters and for cross-lingual transfer, which has the potential to guide the understanding of new methods in this area.

\section{Methodology}

\subsection{Adapter-Based Cross-lingual Transfer}
\label{ss:adapters}
Adapters are components that are inserted into a multilingual pretrained Transformer model (termed the \textit{base model}) to efficiently adapt the large base model to a specific language~\cite{mad-x} or task~\cite{pmlr-v97-houlsby19a}.

The prior adapter-based fine-tuning process typically spans two stages for cross-lingual \emph{task} transfer. First, different language adapters are trained (often separately, per language) with the base model frozen, using the masked language modelling (MLM) objective in a target language. Second, task adapters and a task-specific output head are randomly initialized and inserted into the base model along with now-trained source language (i.e., typically \emph{English}) adapters. In this stage, only the task adapters and a task-specific output head are trainable, while all other parameters are kept fixed. At inference time, the source language adapters are replaced by the language adapter of the target language, while the task adapter is retained, to achieve zero-shot cross-lingual transfer.
In this work, we base our main experiments on the state-of-the-art cross-lingual adapter framework: MAD-X~\cite{mad-x}. See Figure~\ref{fig:adapters} in Appendix~\ref{app:adapter} for architecture details.

\subsection{Gradual Unfreezing and General Scheduled Unfreezing}
\label{ss:su}

First proposed by~\citet{gu}, gradual unfreezing (GU) tunes a subset of parameters of a pretrained model starting from the top layer (see Figure~\ref{fig:my_label}b). Given a model with $L$ layers, assuming that the index $L-1$ refers to the top layer, and interval $k$, GU unfreezes each layer starting from $L-1$ to $0$ in order, every $k$ steps. Once a layer is unfrozen, it remains unfrozen. Hence, the number of trainable parameters increases every $k$ steps under the GU regime.

Let $\textrm{SELECT}(*)$ be a layer-selection function. Let $\textrm{FORWARD}(*)$ be the standard forward pass through all layers, and for a subset $\mathcal{S}$ of layers of the network, let $\textrm{GRADIENT\_UPDATE}(\mathcal{S})$ denote calculating gradients and performing updates on the parameters in $\mathcal{S}$. We define a \emph{generalized scheduled unfreezing} algorithm that encompasses GU, LPFT, and even the recent Surgical fine-tuning method \cite{liang2022surgical}, as well as the other variants we propose later in this work, in Algorithm~\ref{alg:cap}.\footnote{For LPFT, $\textrm{SELECT}(*)$ returns $\emptyset$ for the first $k$ steps and $\mathcal{J} = \{\theta_j\}$ for all layer indices $j$ after the first $k$ steps. In this work, we restrict ourselves to uniform time-intervals and leave the exploration of non-uniform intervals to future work.}

\begin{algorithm}[t]
\scriptsize
\caption{Generalized Scheduled Unfreezing}\label{alg:cap}
\begin{algorithmic}
\Require {An $L$-layer model with layer $j \in \{0, \dots, L-1\}$ parameterized by $\theta_j$. An additional task-specific classification head $C$. Total training budget (steps) $N$. Training interval $k$. Typically $kL \ll N$ for convergence.}
\Statex
\State Initialize $C$, $\theta_j$ for all $j$
\State $\mathcal{S} \gets \{C\}$
\For{$i = 0 \dots \text{N}$}
    \State Sample a data batch $b \sim D$
    \If{$i \mod k == 0$ \textbf{ and } $i \leq kL$}
        \State $\mathcal{J} = \textrm{SELECT}(*)$ \Comment{Set of layer (task adapter) indices to unfreeze.}
        \State $\mathcal{S} \gets \mathcal{S} \cup \{\theta_j : j \in \mathcal{J}\}$
    \EndIf
    \State $\textrm{FORWARD}(*)$
    \State $\textrm{GRADIENT\_UPDATE}(\mathcal{S})$ 
\EndFor
\end{algorithmic}
\end{algorithm}

\subsection{Fisher Information}

We use the \emph{Fisher Information Matrix} ($F$) to investigate changes in learning dynamics. Recent studies have shown that the Fisher Information Matrix correlates well with the generalization capabilities of neural network models~\cite{achille2019time,jastrzebski2021catastrophic}. Conveniently, as a 2nd-order metric based on gradients, $F$ also provides insights into the optimization process. 

In particular, we take the trace of $F$ (i.e., $\Tr(F)$), since the full $F$ is computationally expensive to obtain, and previous work has shown the $\Tr(F)$ correlates well with $F$ and shows similar general trends as the full $F$~\cite{dnncritical}. Let $x$ be data input and consider a network parameterized by weights $w$ that encodes the approximate posterior distribution $p_w(y|x)$. The $\Tr(F)$ is computed using the empirical data distribution $\hat{Q}(x)$ as follows: 

\begin{equation}\label{eqn:fim}
    \begin{aligned}
\Tr(F) = \E_{x\sim \hat{Q}(x)}\E_{\hat{y}\sim p_w(\hat{y}|x)}||\nabla_w \log p_w(\hat{y}|x)||^2
    \end{aligned}
\end{equation}

Note that Eqn.~\eqref{eqn:fim} is not the ``empirical Fisher'' (\citealp{kunstner2019empirical}, empirical Fisher uses the true data label $y$). Hence, one does not need the labels of input data $y$ to calculate the true $F$. They are sampled from the label distribution of the task (i.e., $\hat{y}\sim p_w(\hat{y}|x)$, see pseudo-code in  Appendix~\ref{app:fim_calc}). 

One interpretation of $F$ is that given $w$ and a perturbed version of $w'$ (after applying one step of gradient descent, for example), the KL divergence between $p_w(y|x)$ and $p_{w'}(y|x)$ is given by $\delta w \cdot F \delta w + O(\delta w^3)$ (up to 2nd-order approximation, where  $\delta w$ is the small perturbation in weights)~\cite{naturalgrad}.

$F$ can be considered as a measure of how much a change in weights can affect the network output (i.e., how much information resides in the weights). Intuitively, this means a set of weights with near-zero entries in $F$ likely means they do not significantly affect the network output (and thus task performance). Moreover, $F$ is also a 2nd-order approximation of the Hessian of the loss function~\cite{amari2000infogeo, naturalgrad} and provides information on the curvature of the loss landscape near the current weights, that is, how fast the gradients change during the optimization. 

\section{Experiments}
\subsection{Models}

\sparagraph{Base Models}
The main experiments are conducted with two established pretrained multilingual models: mBERT (base-cased, \citealp{devlin-etal-2019-bert}) and XLM-R (base, \citealp{conneau-etal-2020-unsupervised}).

\noindent \textbf{Adapters.} (\textbf{$^{Ada}$}) We follow the adapter configurations from MAD-X~\cite{mad-x} for cross-lingual transfer, see \S\ref{ss:adapters}. We use pretrained language adapters available in the AdapterHub.\footnote{There are missing language adapters from the AdapterHub for mBERT; we have thus trained our own language adapters following the AdapterHub recommendations for hyperparameter values. Please see Appendix~\ref{app:hparams} for details.
}

\rparagraph{Scheduled Unfreezing Methods}
We apply and analyze two scheduled unfreezing methods from research in other areas of NLP and computer vision to the task of adapter-based cross-lingual transfer.

\noindent\textbf{LPFT} (\citealp{kumar2022finetuning}, \textbf{+LPFT}) first trains the classifier head (linear probing, LP) with the base model frozen, then unfreezes the entire model for fine-tuning (FT). In our setup, we first fine-tune the classifier, then followed by a step of unfreezing all the adapters for fine-tuning. 

\noindent\textbf{Gradual Unfreezing} (\citealp{gu}, \textbf{+GU}) performs top-down unfreezing during training or fine-tuning. We fine-tune with the classifier and the top-most adapter unfrozen, and for every $k$ steps we unfreeze the next adapter and continue.

\subsection{Datasets and Hyperparameters}
We conduct experiments on a diverse set of tasks and target languages. We use MLQA~\cite{lewis-etal-2020-mlqa} and XQuAD~\cite{artetxe-etal-2020-cross-xquad} for question answering (SQuAD, \citealp{squad} for training). We use XNLI~\cite{conneau-etal-2018-xnli} for natural language inference (training on MNLI, \citealp{williams-etal-2018-broad}), and we use XCOPA~\cite{ponti-etal-2020-xcopa} for evaluating causal commonsense reasoning (training on COPA, \citealp{copa}). The data statistics and language codes are summarized in Appendix~\ref{app:data_stats}. We experiment in the zero-shot setting with English-only task data for training and validation.

\rparagraph{Hyperparameters} 
We perform a hyperparameter search with the learning rates of $[$$1$e-$4$, $2$e-$4$, $5$e-$4$, $8$e-$4$$]$ for our main experiments on all datasets except COPA. For COPA, we found a smaller learning rate ($1$e-$5$) is better for scheduled unfreezing methods. See Appendix~\ref{app:hparams} for detailed hyperparameters per task per model. For scheduled unfreezing experiments, we search for the hyperparameter $k$ in the following range [25, 50, 100, 800, 1000]. The reported results are averaged across 4 runs on A100 or V100 GPUs.

\section{Results and Analysis}

\subsection{Scheduled Unfreezing Improves Cross-Lingual Generalization}

\begin{figure*}
    \centering
    \includegraphics[width=0.9\textwidth]{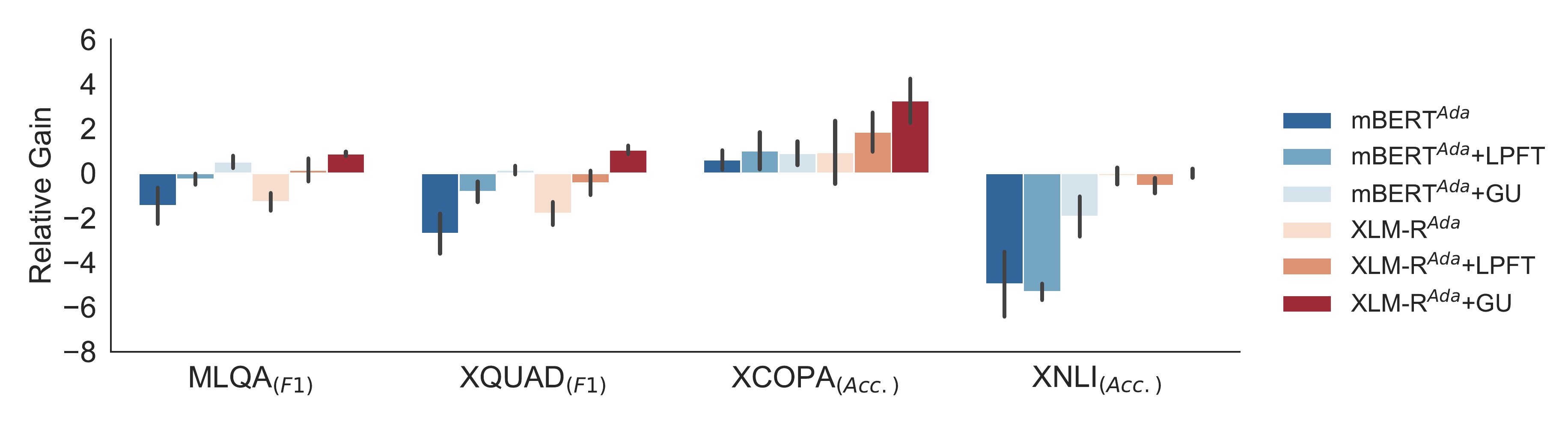}
    \vspace{-2mm}
    \caption{The relative performance of adapters fine-tuned with scheduled unfreezing (i.e., GU-based and LPFT-based task adapters) and standard fine-tuned task adapters with full fine-tuning of mBERT and XLM-R.}
    \label{fig:rel_gain}
    \vspace{-1.5mm}
\end{figure*}

Figure~\ref{fig:rel_gain} shows the relative performance of (a) task adapters fine-tuned in the standard way ($^{Ada}$), (b) GU- ($^{Ada}$+GU) and LPFT-tuned adapters ($^{Ada}$+LPFT) compared to full fine-tuning with mBERT and XLM-R on all datasets.\footnote{Baseline full parameter fine-tuning results used for Figure~\ref{fig:rel_gain} are in Table~\ref{tab:baselines} of the Appendix.} We report the results averaged across all target languages in the respective datasets. Our experiments show that both LPFT and GU are effective in closing the gap to full fine-tuning across all tasks and models. Moreover, GU-trained task adapters perform better, even exceeding the performance of full fine-tuning in some cases.\footnote{Although we arrived at a different conclusion from \citet{raffel2022t5}, we emphasize that \citet{raffel2022t5} compared full fine-tuning + GU with standard full fine-tuning, which is different from our work. We evaluate the cross-lingual transfer setup, which inherently comes with distribution shifts. }

Our results suggest that scheduled unfreezing can do more than just mitigate catastrophic forgetting. Even in a CF-free setting like ours, they achieve better generalization for cross-lingual transfer. We focus on GU as the scheduled unfreezing method for further analyses, since it produced better empirical results than LPFT. 

Since the training data for both XQuAD and XNLI are well over 50k instances; this amount of annotated task data might be unrealistic for many tasks in practice, even in English. We simulated low-data training settings (details and results are in Appendix~\ref{app:more_exp}, Table~\ref{tab:lessshot}), and found that even with a smaller amount of training data, we still observe the advantages of GU over standard task adapter fine-tuning.

\subsection{Scheduled Unfreezing Beyond Mitigating Catastrophic Forgetting}

To understand why scheduled unfreezing helps even in the CF-free setting, we examine the learning dynamics during the training of task adapters. 

Due to the unfreezing of task adapters at different times, the model has access to a different number of trainable parameters, which affects the optimization and information encoding for adapters differently under scheduled unfreezing than in standard fine-tuning. Hence, we draw our attention to higher-order metrics, captured in \emph{$\Tr(F)$}.

\begin{figure*}[t!]
    \centering
    \begin{subfigure}[t]{0.29\textwidth}
        \centering
        \includegraphics[clip, width=0.92\textwidth]{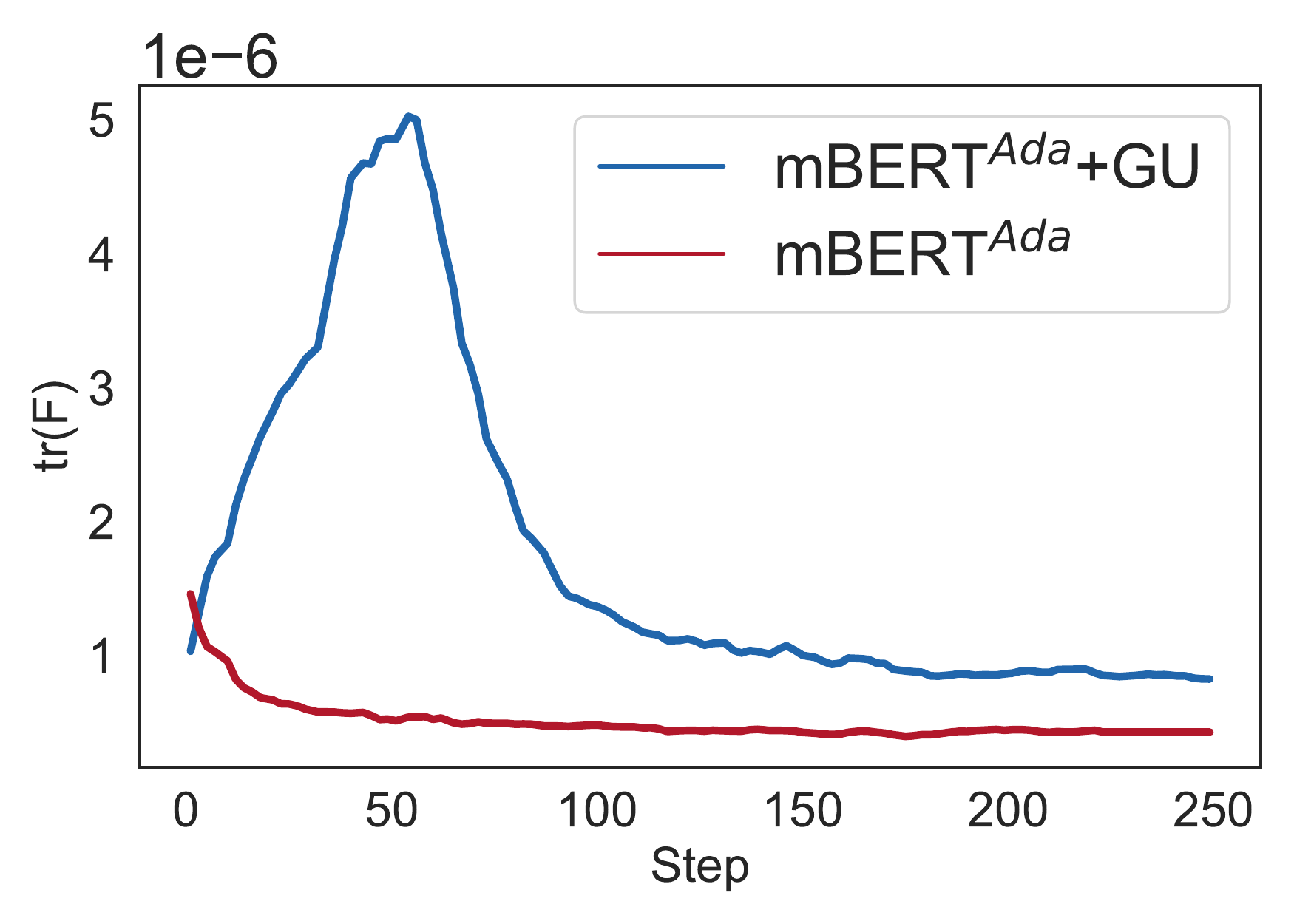}
        \caption{mBERT SQuAD}%
        \label{fig:a} 
    \end{subfigure}
    \hfill
    \begin{subfigure}[t]{0.29\textwidth}
        \centering
        \includegraphics[clip,width=0.92\textwidth]{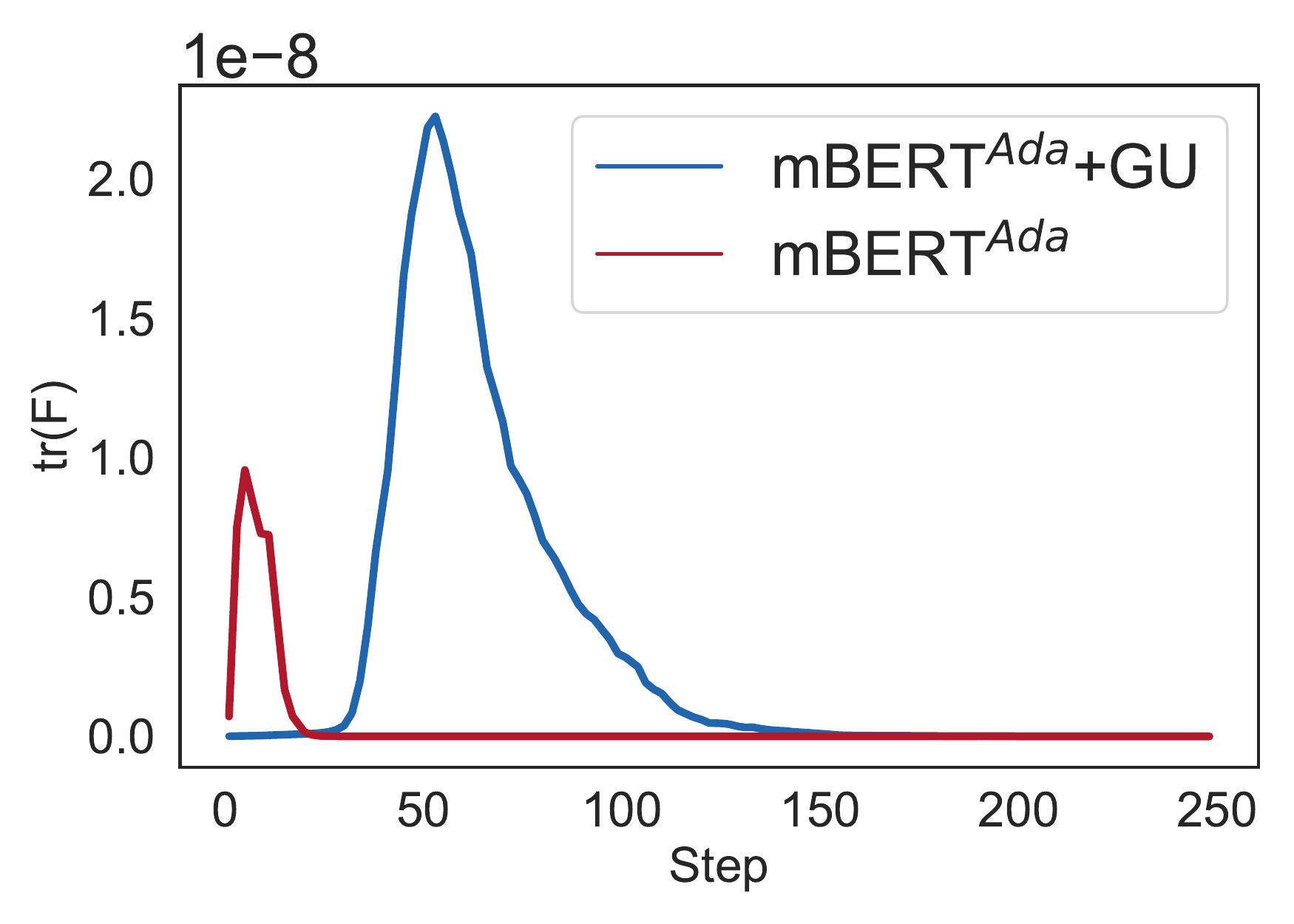}
        \caption{mBERT COPA}%
        \label{fig:b}
    \end{subfigure}
    \hfill
    \begin{subfigure}[t]{0.29\textwidth}
        \centering
        \includegraphics[clip, width=0.92\textwidth]{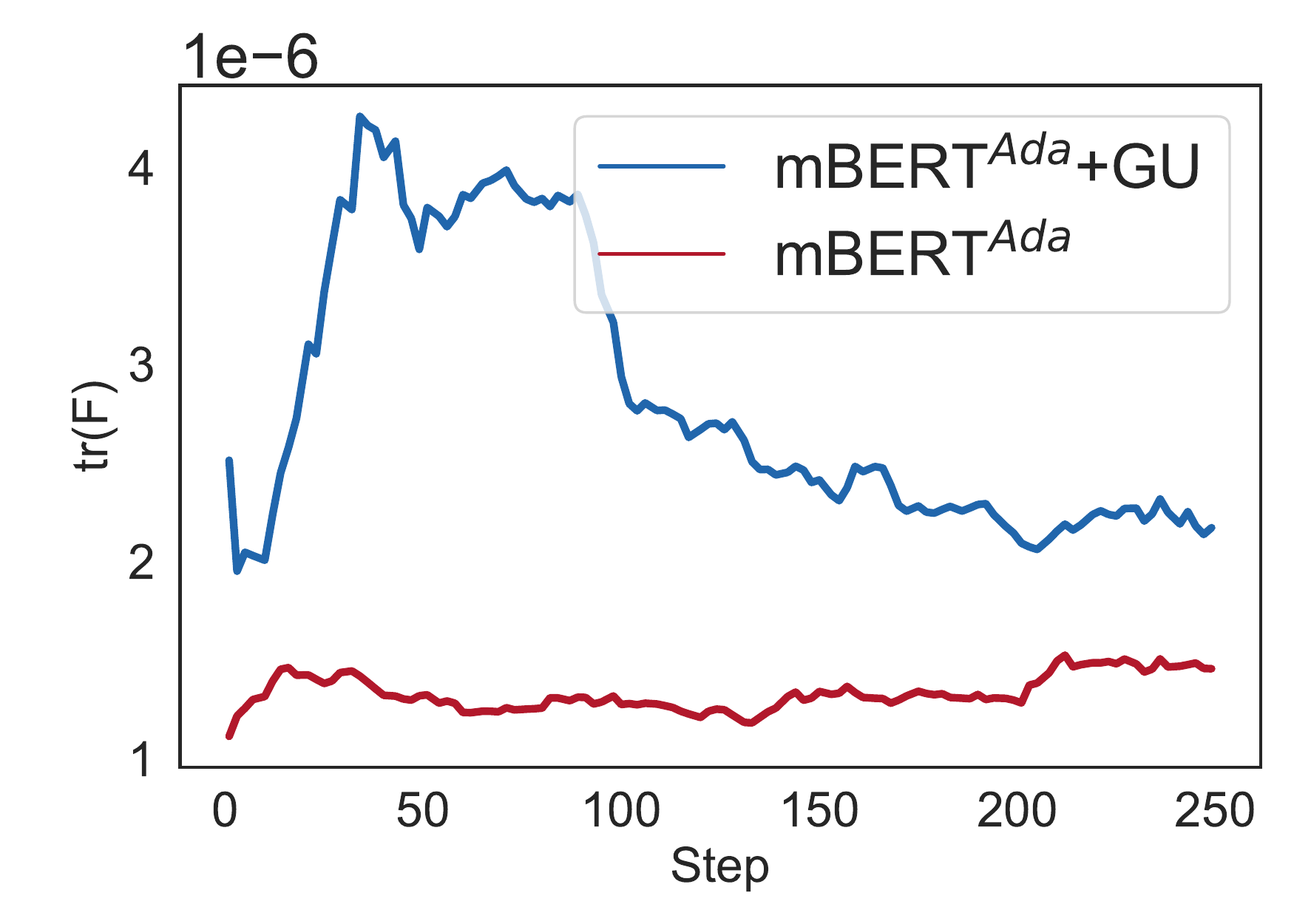}
        \caption{mBERT MNLI}%
        \label{fig:c}
    \end{subfigure}
    \medskip
    \begin{subfigure}[t]{0.29\textwidth}
        \centering
        \includegraphics[clip, width=0.92\textwidth]{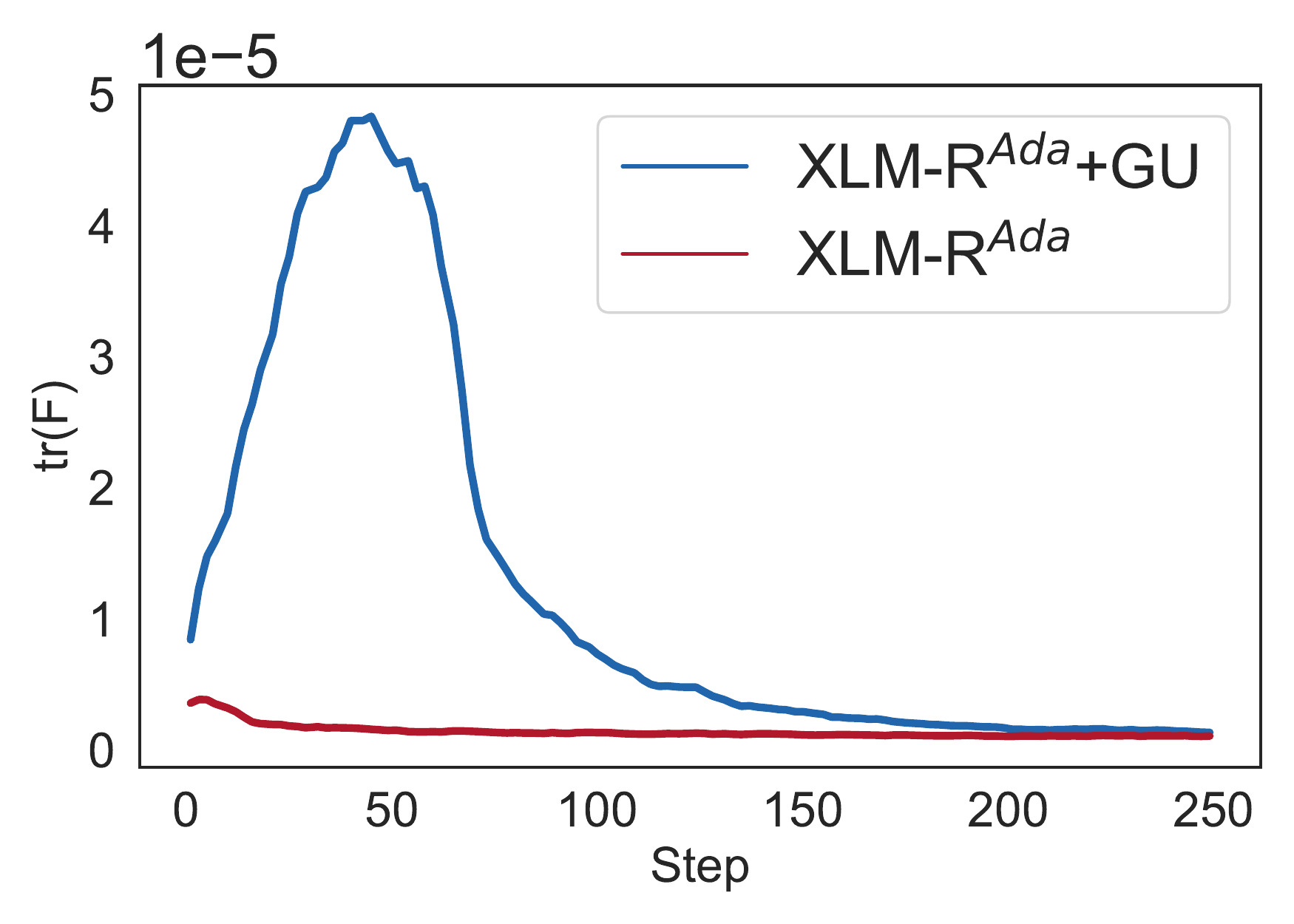}
        \caption{XLM-R SQuAD}%
        \label{fig:d} 
    \end{subfigure}
    \hfill
    \begin{subfigure}[t]{0.29\textwidth}
        \centering
        \includegraphics[clip,width=0.92\textwidth]{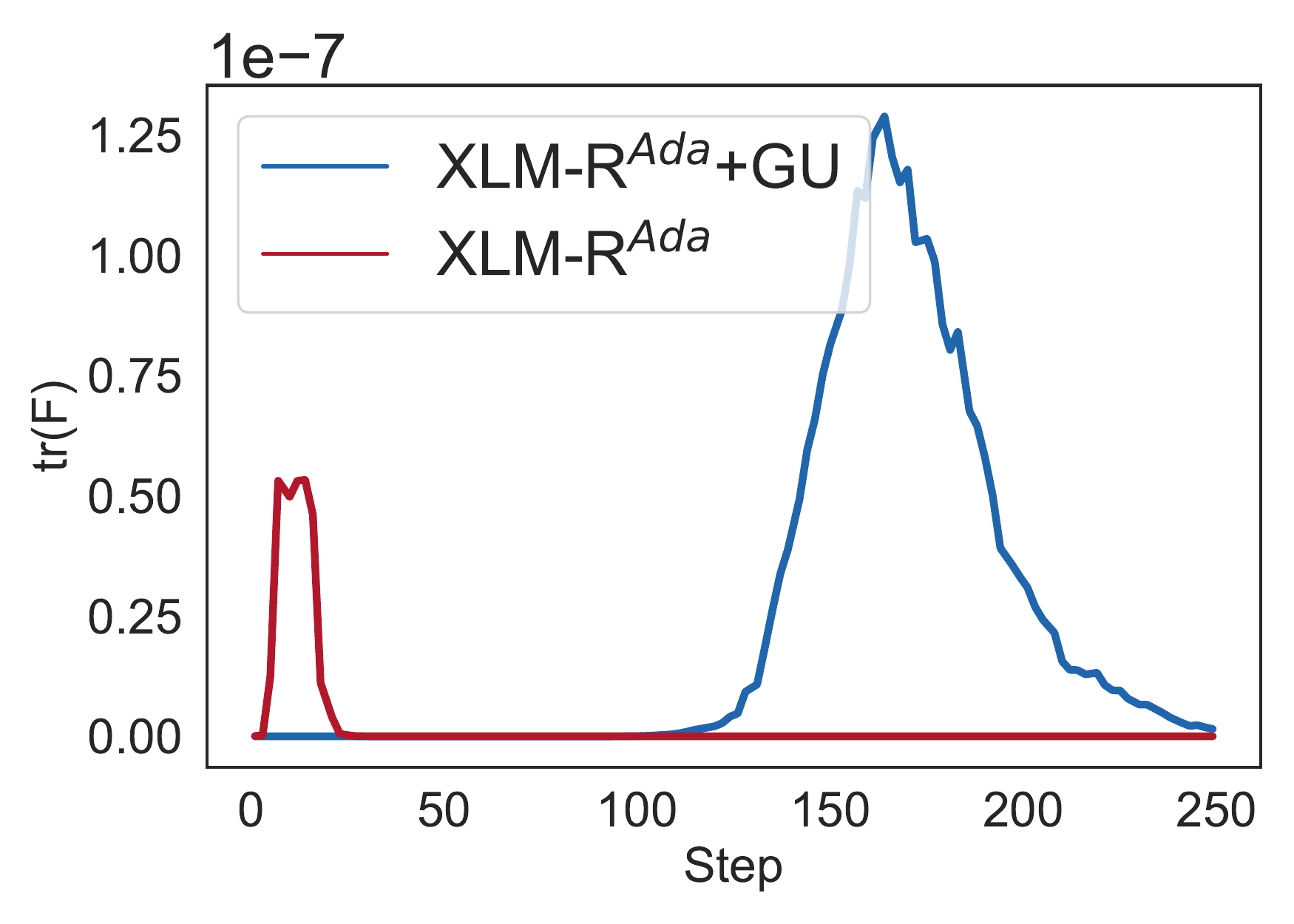}
        \caption{XLM-R COPA}%
        \label{fig:e}
    \end{subfigure}
    \hfill
    \begin{subfigure}[t]{0.29\textwidth}
        \centering
        \includegraphics[clip, width=0.92\textwidth]{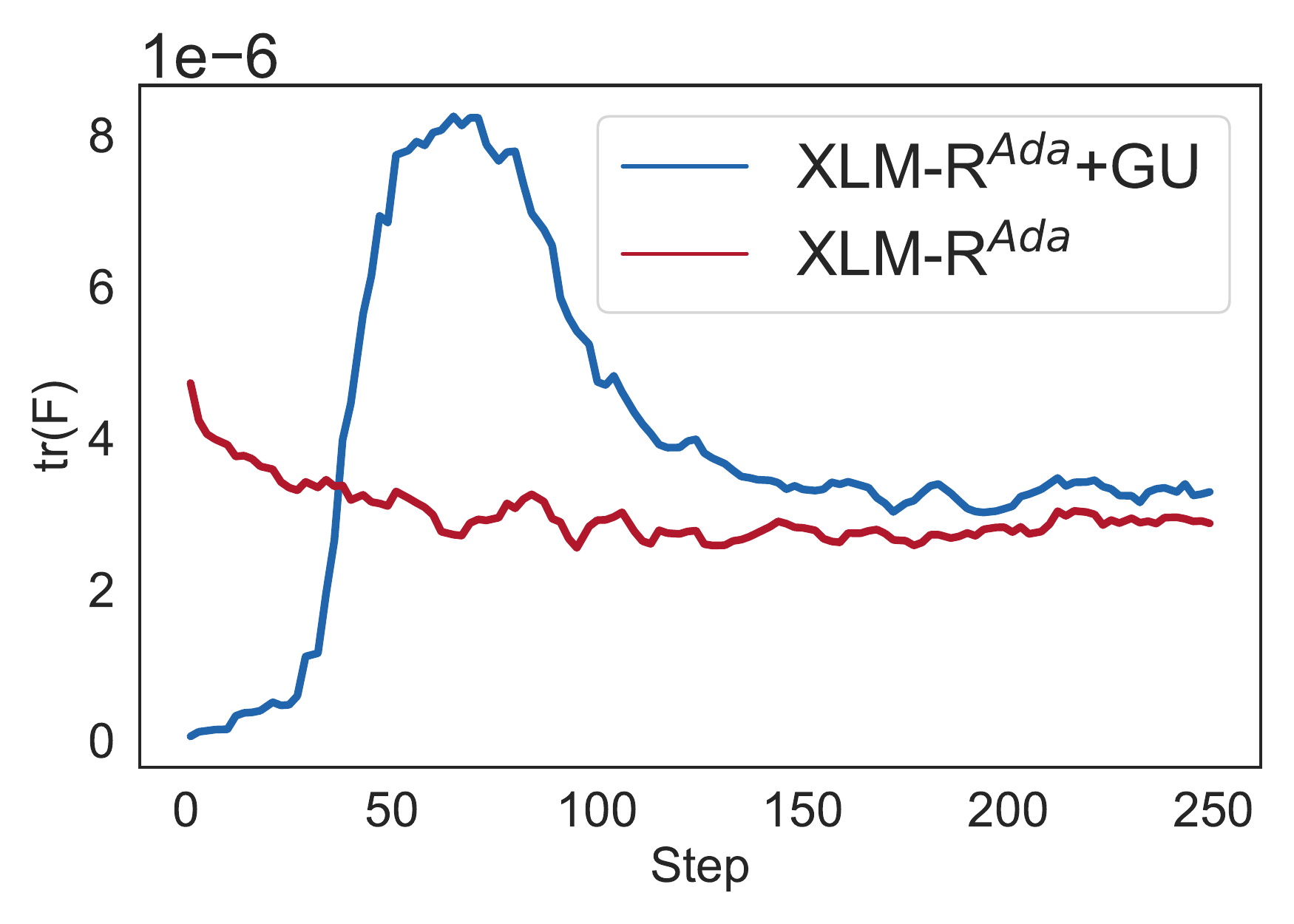}
        \caption{XLM-R MNLI}%
        \label{fig:f}
    \end{subfigure}
    \vspace{-1.0mm}
    \caption{Average $\Tr(F)$ per adapter during standard training versus using gradual unfreezing. Every point on the horizontal axis is 100 training steps for all datasets (except for XCOPA which is 50 steps).}
    \label{fig:fish}
    \vspace{-1.5mm}
\end{figure*}

\rparagraph{GU Changes Learning Dynamics} We plot $\Tr(F)$ (moving average, normalized by the number of trainable adapters at the given step) during optimization. Figure~\ref{fig:fish} shows $\Tr(F)$ during training on three datasets.\footnote{We calculate $\Tr(F)$ every 100 steps for all datasets (except every 50 steps for XCOPA due to its small training size).} The plots indicate that GU significantly changes the learning dynamics of adapters compared to their standard fine-tuning. With GU, due to the model having fewer parameters to encode the same amount of data initially, the $\Tr(F)$ curve is higher than in standard fine-tuning. The training process induces a $\Tr(F)$ curve that has a distinctive ``hill'' shape (i.e., a ``learning period", with fast changes of gradients, and hence weights).

\rparagraph{Effects of the Unfreezing Schedule on $\Tr(F)$ and Generalization}\label{sec:fim_gen}
While we have empirically validated that scheduled unfreezing changes $\Tr(F)$ during learning, it is unclear which factors are the main drivers that impact $\Tr(F)$, and potentially improve generalization. Previous work has studied factors such as learning rate, weight decay, optimizer and loss functions~\cite{achille2019time,jastrzebski2021catastrophic}. However, based on this work, another novel relevant factor is the \textit{unfreezing schedule} (i.e., the number of parameters to update at each optimization step). Here, we aim to further understand how unfreezing schedules change the learning dynamics and generalization. 

We found that the sensitivity of the learning dynamics to schedules depends on the dataset and the base model. Hence, we focus on XLM-R for MLQA/XQuAD and mBERT for XNLI. These two settings are the most sensitive to schedules and best illustrate the effects, but they are different in terms of models and tasks to examine general patterns.

\begin{figure}
        \centering
    \begin{subfigure}[t]{0.98\linewidth}
        \centering
        \includegraphics[clip, width=0.8\textwidth]{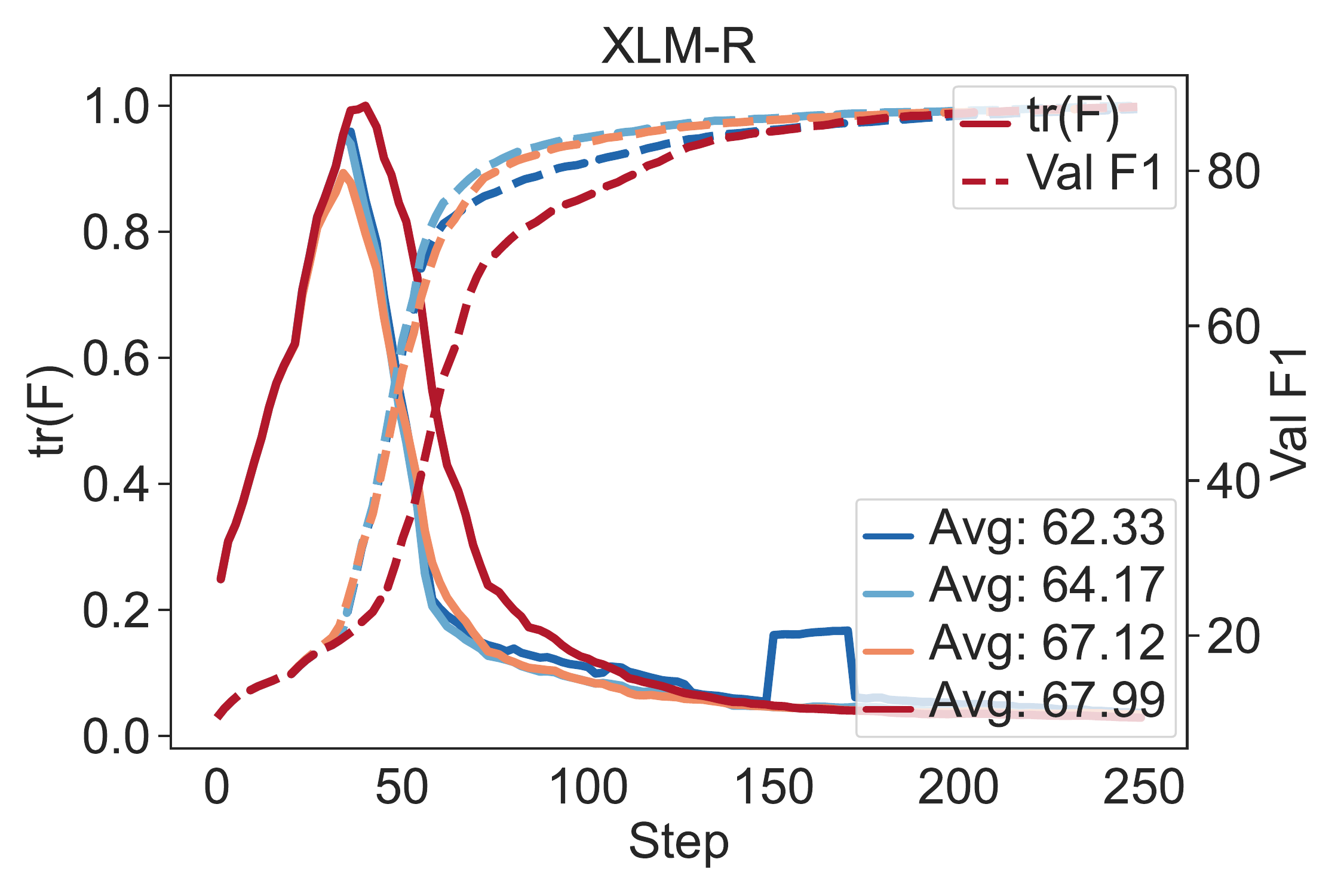}
        \caption{SQuAD}%
        \label{fig:schedule_xlmr} 
    \end{subfigure}
    \medskip

    \begin{subfigure}[t]{0.98\linewidth}
        \centering
        \includegraphics[clip, width=0.8\textwidth]{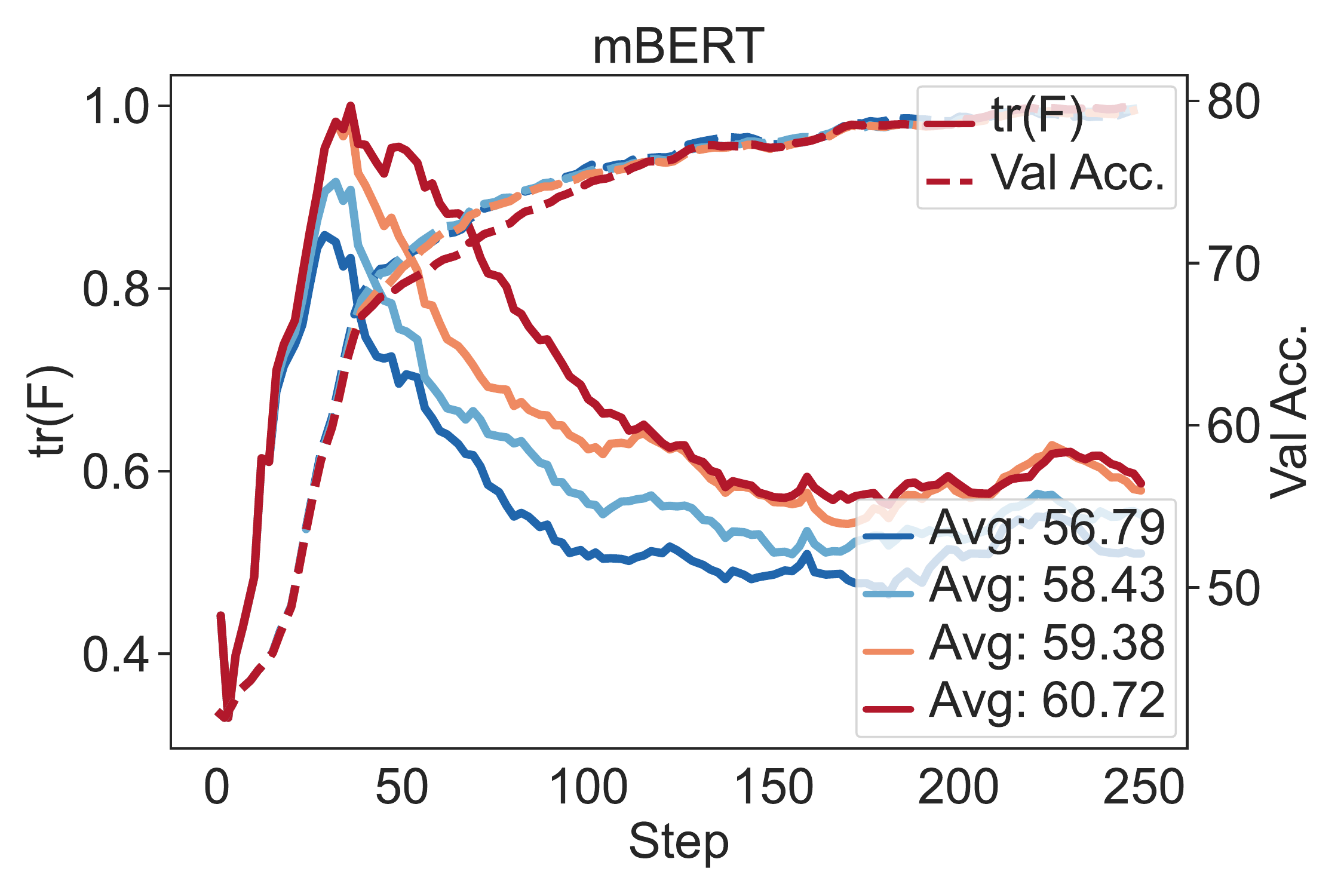}
        \caption{MNLI}%
        \label{fig:schedule_mbert} 
    \end{subfigure}
    \caption{Average $\Tr(F)$ per adapter (normalized between 0-1 to plot together with the validation curve) and validation F1 or accuracy using a randomly sampled schedule. The average results indicated in the legend are the \emph{averaged cross-lingual} transfer results. a) averaged F1 of MLQA and XQuAD, b) XNLI.}
    \label{fig:schedule_hl} 
\end{figure}

We randomly sampled 9 schedules (which are effectively permutations of layer indices, where we also treat the standard top-down order as the 10th schedule) that start unfreezing at either layer 10 or 9. The remaining experimental conditions are unchanged.\footnote{We consider these top layers, which likely carry similar information, in contrast to lower layers such as 0 or 1. The 11th layer is always unfrozen at the beginning, and we reject permutations of layer indices that do not start with 10 or 9. An accepted sampled schedule could look like [10, 1,  0, 3,  9,  6,  8, 2,  5,  7,  4].} We then aggregate runs with similar cross-lingual transfer results.\footnote{This aggregation is done by: (i) sorting the cross-lingual transfer results, then (ii) taking the largest `ungrouped' value, (iii) finding runs that are within 1 point of performance delta, and then (iv) grouping them.} 

We plot the $\Tr(F)$ along with the validation metrics in Figure~\ref{fig:schedule_hl}. We observe that decreases in $\Tr(F)$ from the peak value are associated with rapid generalization (cf., a drastic increase of the validation metrics) of the network. Previous work points out that the peak of $\Tr(F)$ -- with contradicting claims from different works~\cite{achille2019time, dnncritical, jastrzebski2021catastrophic} -- correlates or anti-correlates with generalization, which indicates the underlining relationship may be more complex than just the peak value of the $\Tr(F)$ curve. 

Indeed, Figure~\ref{fig:schedule_hl} shows that a \emph{wider} $\Tr(F)$ curve (a longer learning period) often accompanies a large peak $\Tr(F)$ value during the early phase of learning, and leads to better cross-lingual generalization during the test time. This could potentially lead to an additional new avenue of manipulating optimization with a regularization loss for cross-lingual transfer, which we leave for future work.

To the best of our knowledge, this is the first evidence indicating that $\Tr(F)$ correlates with cross-lingual transfer performance in parameter-efficient fine-tuning and text-based Transformers. These results suggest that early-phase learning dynamics affect the generalization cross-lingually later on, and $\Tr(F)$ can be a potential measurement to study cross-lingual generalization. From the above results, we conjecture that \emph{inducing large $\Tr(F)$ and a longer learning period would improve generalization in the CF-free setting},\footnote{Although we empirically observe similar patterns as \citet{jastrzebski2021catastrophic}, we interpret the results differently. The prior work argues that a sub-optimal small learning rate induces a higher peak $\Tr(F)$ during learning (termed catastrophic Fisher explosion), and regularizing $\Tr(F)$ to a smaller value helps with learning and generalization. In this work, we study cross-lingual generalization, which is a \textbf{shifted distribution} at test time instead of in-distribution generalization (the focus of the prior work). Furthermore, we also use a different optimizer, large learning rates and work in a CF-free setting. More discussion on combining with a regularizer is given in Appendix~\ref{app:discussion}.} which motivates our experiments in the next section.

\begin{table*}[h]
\setlength\tabcolsep{2pt}
    \centering
    \resizebox{0.93\textwidth}{!}{
    \tiny
    \begin{tabular}{l| ccc| ccc }
    \toprule
    \multicolumn{1}{c}{} & \multicolumn{3}{c}{\textit{MLQA (F1 / EM)}} & \multicolumn{3}{c}{\textit{XQuAD (F1 / EM)}} \\
    \midrule
        \textbf{Method} & En & Lowest (Ar) & \textbf{Average} & 
        En & Lowest (Th) & \textbf{Average} \\
    mBERT$^{Ada}$ & 78.99/65.85	& 45.76/28.77	&  55.40$\pm$0.94 / 37.07$\pm$0.72 & 83.58/71.74 & 34.53/38.89	& 60.63$\pm$1.04 / 43.90$\pm$0.85 \\
    mBERT$^{Ada}$+Rand & 79.22/65.94	& 46.65/29.84		& 55.93$\pm$0.21 / 37.54$\pm$0.31 & 83.86/72.31	&
    38.91/38.72	& 61.68$\pm$0.33 / \underline{47.42}$\pm$0.55  \\
    mBERT$^{Ada}$+GU & 78.04/64.20 & 47.96/29.30	& \gbc\textbf{57.37}$\pm$0.32 / \underline{38.27}$\pm$0.27 & 83.21/71.55	& 44.08/35.46	& \gbc\textbf{63.48}$\pm$0.22 / 46.76$\pm$0.44\\
    mBERT$^{Ada}$+FUN & 78.82/65.29	& 48.20/30.90	&\gbc\underline{57.33}$\pm$0.51 / \textbf{38.29}$\pm$0.63 & 83.71/71.83	& 42.55/42.84	& \gbc\underline{63.25}$\pm$0.26 / \textbf{49.09}$\pm$0.48 \\
    \midrule
        \textbf{Method} & En & Lowest (Ar) & \textbf{Average} & 
        En & Lowest (Ar) & \textbf{Average} 
        \\
    XLM-R$^{Ada}$ & 79.52/65.99 & 51.74/33.33  & 61.31$\pm$0.46 / 42.10$\pm$0.42 & 83.48/72.69	& 65.47/48.89	& 70.09$\pm$0.60 / 53.77$\pm$0.40\\
    XLM-R$^{Ada}$+Rand & 80.32/67.01	& 50.33/36.29	& 61.36$\pm$1.69 / 41.59$\pm$1.96& 84.76/73.74	& 63.69/46.30	& 69.99$\pm$1.47 / 52.06$\pm$2.04\\

    XLM-R$^{Ada}$+GU & 80.37/66.77	&  55.16/35.49	& \gbc\textbf{63.47}$\pm$0.12 / \textbf{43.55}$\pm$0.11 & 84.49/73.57	& 67.83/50.80	& \gbc\textbf{73.04}$\pm$0.22 / \textbf{55.93}$\pm$0.15 \\
    XLM-R$^{Ada}$+FUN & 80.92/66.70	& 53.17/37.73	& \gbc\underline{63.10}$\pm$0.79 / \underline{43.37}$\pm$0.51& 84.91/73.80 & 66.69/49.24& \gbc\underline{72.34}$\pm$0.40 / \underline{55.21}$\pm$0.63\\
   \midrule
    \multicolumn{1}{c}{} & \multicolumn{3}{c}{\textit{XCOPA (Accuracy)}} & \multicolumn{3}{c}{\textit{XNLI (Accuracy)}} \\
    \midrule
   \textbf{Method} & 
        En & Lowest (It) & \textbf{Average} & 
        En & Lowest (Sw) & \textbf{Average} 
        \\
        mBERT$^{Ada}$ & 63.80	&  50.16 & \underline{53.99}$\pm$0.49 & 82.05	& 37.45	& 57.78$\pm$1.68 \\
    mBERT$^{Ada}$+Rand & 65.00& 50.96	& 53.84$\pm$0.71 & 81.64	& 45.95	& 59.87$\pm$0.96 \\
    mBERT$^{Ada}$+GU & 66.60	& 50.00	& \gbc\textbf{54.29}$\pm$0.60 & 81.79	& 54.06		& \gbc\textbf{61.67}$\pm$1.04 \\
    mBERT$^{Ada}$+FUN & 66.40& 50.68	& \gbc\underline{53.98}$\pm$0.64 & 81.70	& 53.73		& \gbc\underline{61.36}$\pm$0.51\\

    \midrule
   \textbf{Method} & 
        En & Lowest (Ht) & \textbf{Average} & 
        En & Lowest (Sw) & \textbf{Average} 
        \\
    XLM-R$^{Ada}$ & 65.20	& 51.28	& 55.93$\pm$1.58 & 84.31	& 68.16	& \underline{73.31}$\pm$0.44 \\
    XLM-R$^{Ada}$+Rand & 67.20	& 52.12	& 57.05$\pm$0.42 & 84.52	&
    67.29	& 72.68$\pm$0.56\\
    XLM-R$^{Ada}$+GU &  66.00	& 52.52 & \gbc\textbf{58.24}$\pm$1.11 & 84.24	& 68.24 & \gbc\textbf{73.44}$\pm$0.24  \\
    
    XLM-R$^{Ada}$+FUN & 67.80 &  52.08	& \gbc\underline{58.11}$\pm$0.94 &  84.72	& 67.48 & \gbc73.13$\pm$0.53 \\

    \bottomrule
    \end{tabular}
    }
    \vspace{-0.5mm}
    \caption{Zero-shot transfer results across four datasets: MLQA, XQuAD, XCOPA and XNLI. Average is the cross-lingual average without English. We bold the highest and underline the second-highest average value. \textit{Lowest} denotes the task performance for the lowest-performing target language per each evaluation dataset and base model.}
    \label{tab:main_all}
    \vspace{-2mm}
\end{table*}

\subsection{Auto-Scheduling by $\Tr(F)$}

We have observed that scheduled unfreezing effectively changes the learning dynamics of the task adapters. A natural follow-up question is: \textit{If freezing certain parameters changes the learning dynamics, can we systematically and automatically select which task adapter to unfreeze next?} 

To answer this question, we propose to select the next layer to unfreeze based on ranked $\Tr(F)$ during training (i.e., Figure~\ref{fig:my_label}c). According to our hypothesis, an induced large and wide $\Tr(F)$ during learning leads to better generalization (discussed in Section~\ref{sec:fim_gen}).\footnote{$\Tr(F)$ is calculated every $k$ steps for $L$ times, on $40$ batches (maximum) of training data. We consider the additional computational cost to be negligible compared to GU. 
}

\begin{figure*}[h!]
    \centering
    \begin{subfigure}[t]{0.32\textwidth}
        \centering
        \includegraphics[clip, width=0.9\textwidth]{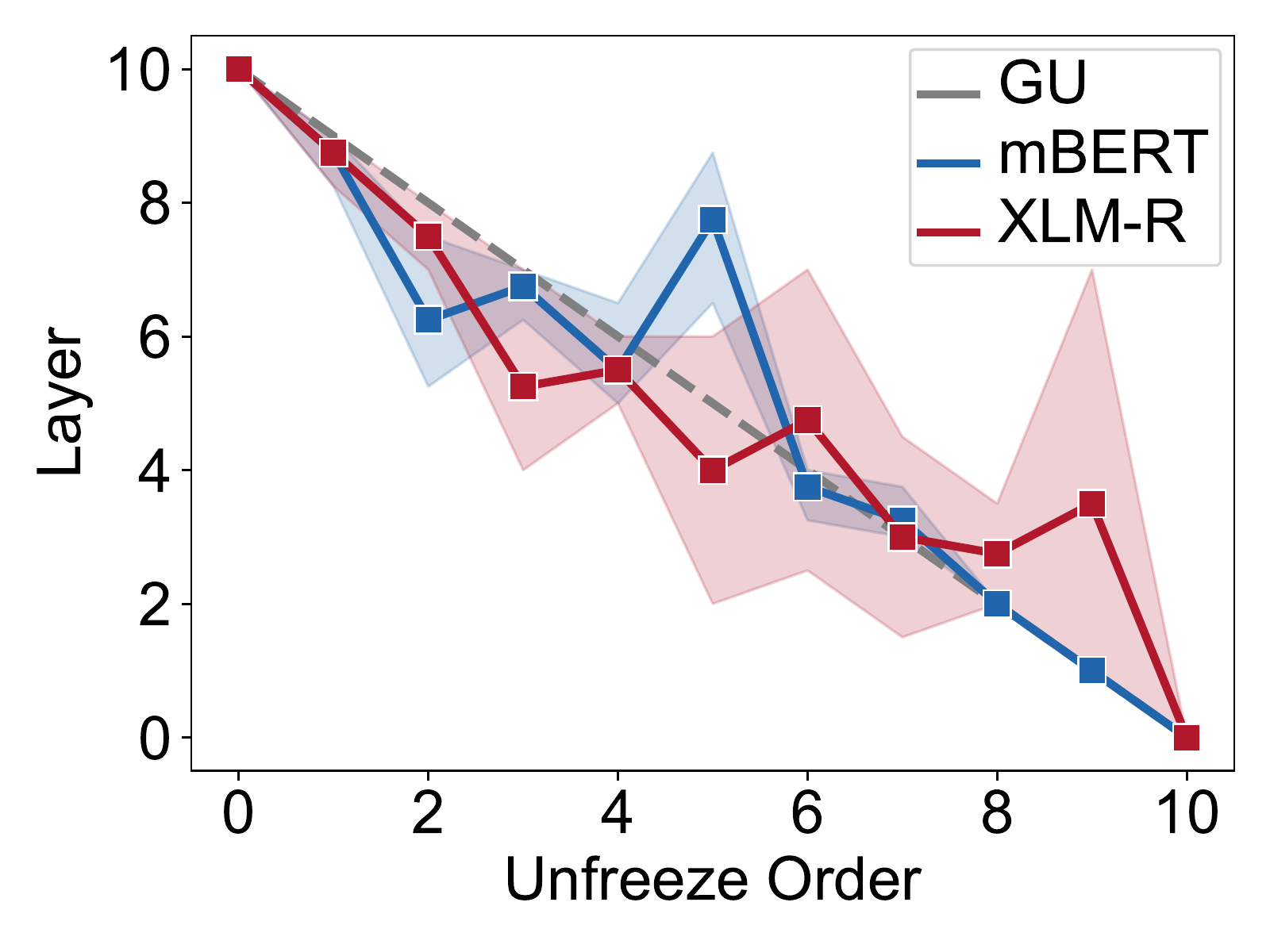}
        \caption{SQuAD}%
        \label{fig:d} 
    \end{subfigure}
    \hfill
    \begin{subfigure}[t]{0.32\textwidth}
        \centering
        \includegraphics[clip,width=0.9\textwidth]{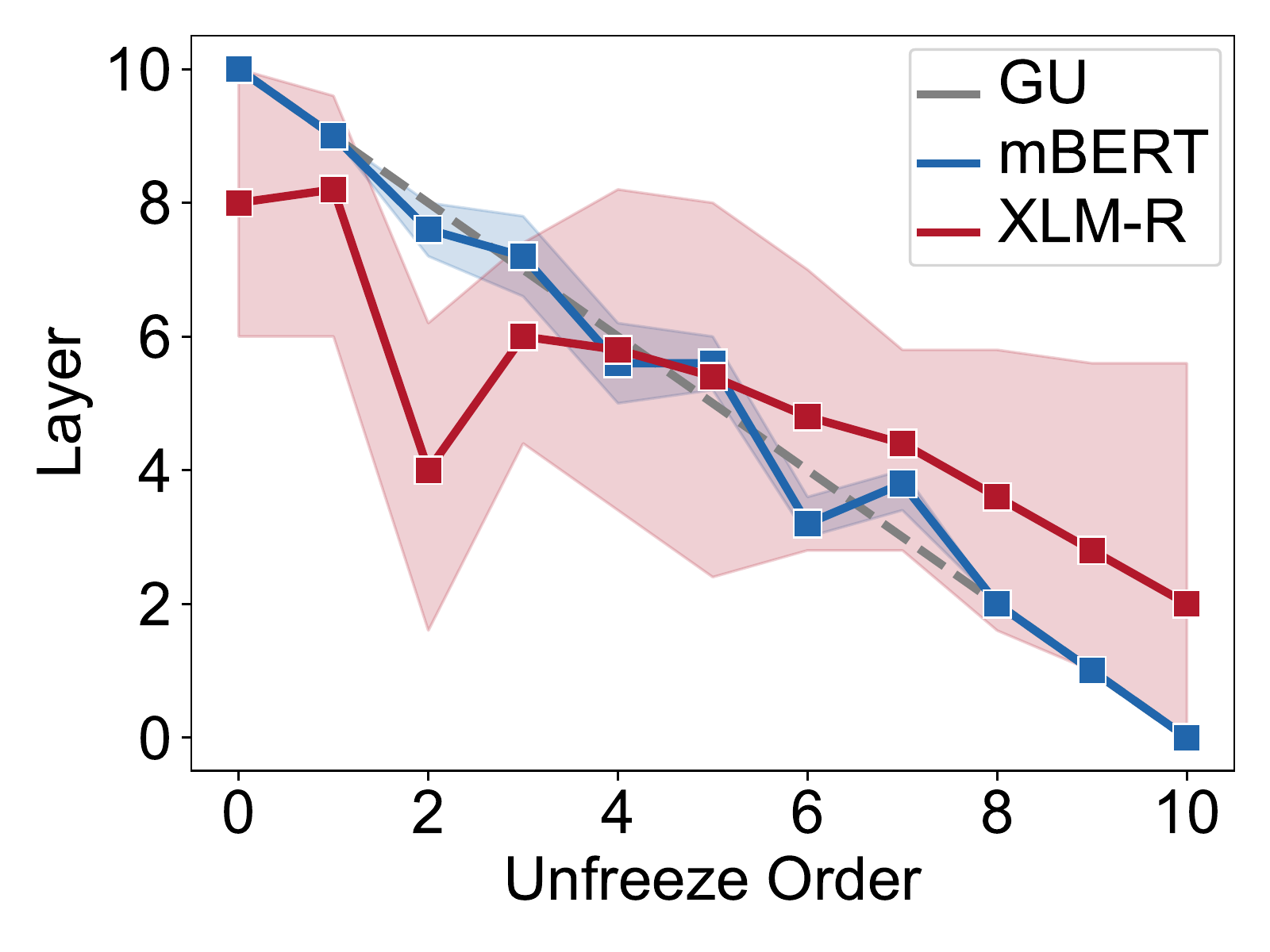}
        \caption{COPA}%
        \label{fig:e}
    \end{subfigure}
    \hfill
    \begin{subfigure}[t]{0.32\textwidth}
        \centering
        \includegraphics[clip, width=0.9\textwidth]{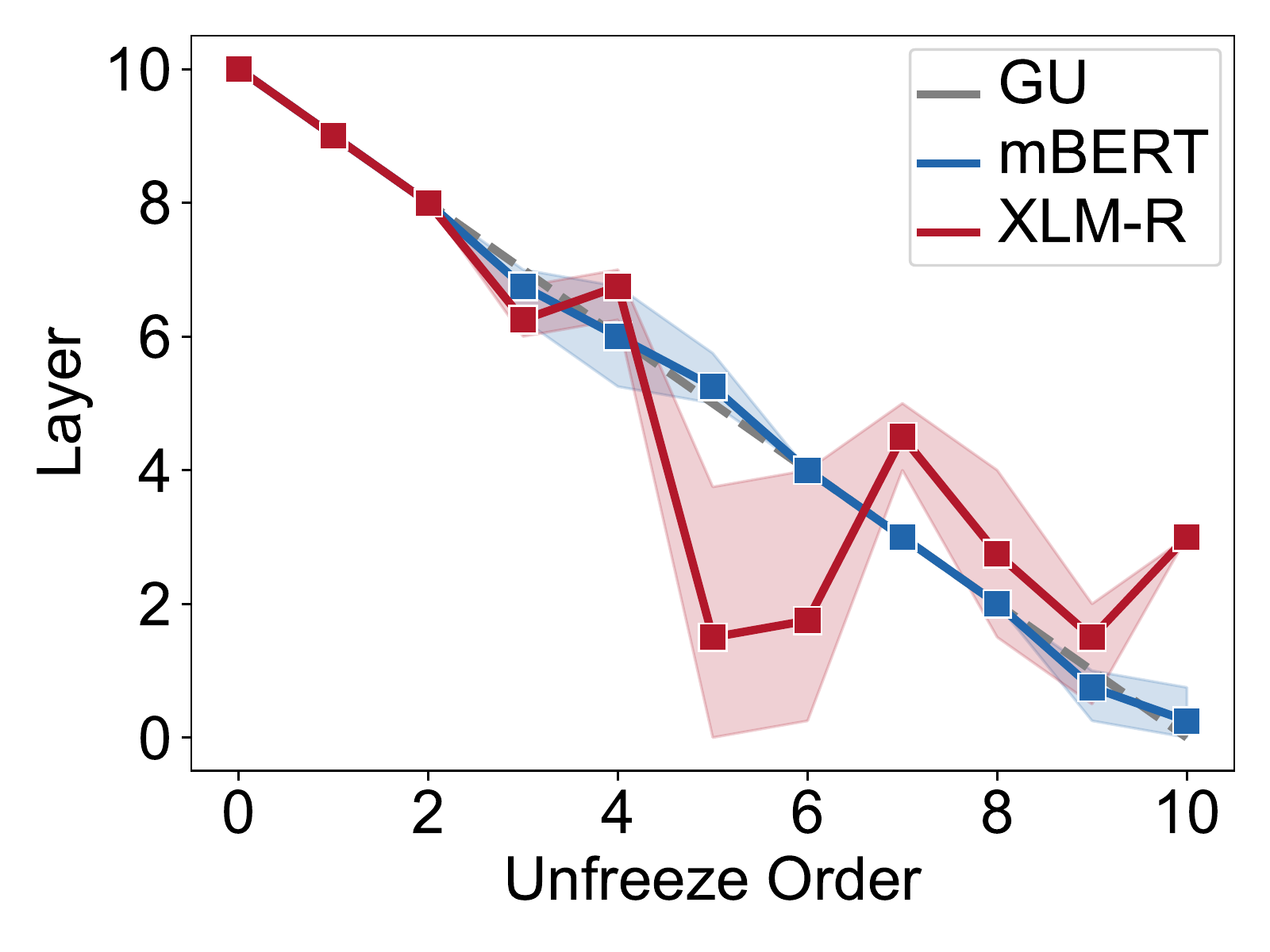}
        \caption{MNLI}%
        \label{fig:f}
    \end{subfigure}
    \vspace{-1.0mm}
    \caption{Averaged unfreezing schedules for GU and FUN with different base models.}
    \label{fig:fim_order}
    \vspace{-1.5mm}
\end{figure*}

\rparagraph{$\Tr(F)$-based Scheduled Unfreezing (FUN) Recovers GU} Surprisingly, the $\Tr(F)$-based schedules recover the transfer performance as well as the top-down heuristic schedule (i.e., GU) in many cases. To illustrate this, we plot the unfreezing schedules generated by FUN along with the top-down schedule of GU (diagonal line) for all our experiments in Figure~\ref{fig:fim_order}. From the figure, we can see that FUN nearly perfectly recovers the top-down schedule of GU for mBERT. We conjecture that GU is implicitly maximizing $\Tr(F)$ at every unfreezing step. The XLM-R-based experiments largely follow the top-down order with more variance, which is likely due to noise in $\Tr(F)$ estimation (for efficiency, we don't use the entire training data for estimation). 

We show the results across all datasets in Table~\ref{tab:main_all} and include an additional baseline (\textit{+Rand}) that randomly selects the next layer to unfreeze at every time interval $k$ (see Appendix~\ref{app:full_results} for detailed results).  
Table~\ref{tab:main_all} shows that the FUN scheduler achieves near-identical results as GU with the mBERT model, which matches the observations in Figure~\ref{fig:fim_order}. We also achieve comparable results as GU with XLM-R. The results are well above the random unfreezing baselines and standard training (e.g., improving mBERT from standard training on XNLI for 3.58 points, or the average of 2.03 points across all 4 datasets etc.). Although the source English performance is not the focus of our work, we also find that FUN achieves better English results in most of the cases (e.g., XLM-R with FUN is better than GU in English for 6 out of 8 cases, which shows the potential of FUN beyond the context of cross-lingual transfer explored here).

In addition, we highlight that scheduled unfreezing improved transfer results for the lowest-performing language across the board. For example, FUN improved the lowest-performing language under standard adapter training in all 4 cases for the mBERT model, and in 3 out of 4 cases for XLM-R. Some gains are quite substantial, e.g., Thai with mBERT increased from 34.53 to 42.55 (see Table~\ref{tab:main_all}, the \textit{Lowest} column for XQuAD).

\rparagraph{Beyond MAD-X}
Although less studied in the cross-lingual transfer setting, to show the generality of our findings, we perform additional experiments with LoRA adapters~\cite[see the original paper for details]{hu2022lora}. From Table~\ref{tab:lora}, we observe that GU and FUN have comparable performance on average (FUN is better than GU for MLQA and GU is better than FUN for XQuAD with mBERT, and comparable for XLM-R) for cross-lingual transfer. More importantly, both unfreezing methods consistently outperform the standard tuning of LoRA. Additional results for other tasks with LoRA can be found in Table~\ref{tab:app_lora} in Appendix~\ref{app:lora}, which also demonstrates comparable results for GU and FUN. Our results indicate that scheduled unfreezing is a general method applicable to different adapter types and improves generalization.

\begin{table}[h]
\setlength\tabcolsep{2pt}
    \resizebox{0.98\linewidth}{!}{
    \centering
    \tiny
    \begin{tabular}{l| ccc }
    \toprule
    \multicolumn{1}{c}{} & \multicolumn{3}{c}{\textit{XQuAD (F1 / EM)}} \\
    \midrule
    \textbf{Method} & En & Lowest (Th) & \textbf{Average} \\
    mBERT$^{LoRA}$ & 83.68/71.50	& 37.56/29.44	&  61.53$\pm$0.37 / 45.01$\pm$0.44  \\
    mBERT$^{LoRA}$+GU & 84.24/72.75 & 40.73/31.41	& \gbc\textbf{63.12}$\pm$0.20 / \textbf{45.96}$\pm$0.38 \\
    mBERT$^{LoRA}$+FUN & 83.58/71.83	& 39.88/28.59	&\gbc\underline{62.92}$\pm$0.46 / \underline{45.80}$\pm$0.42 \\
    \midrule

    XLM-R$^{LoRA}$ &  83.35/72.27	& 58.09/42.10 	& 68.92$\pm$1.16 / 51.50$\pm$1.53  \\
    
    XLM-R$^{LoRA}$+GU & 84.70/73.12 & 66.72/49.50 & \gbc\textbf{72.27}$\pm$0.12 / \textbf{54.67}$\pm$0.34  \\
    
    XLM-R$^{LoRA}$+FUN & 
     84.22/72.29 & 65.69/48.62 & \gbc\underline{72.13}$\pm$0.18 / \underline{54.53}$\pm$0.14 
    \\
   
   \midrule   
   \multicolumn{1}{c}{} & \multicolumn{3}{c}{\textit{MLQA (F1 / EM)}} \\
    \midrule
    \textbf{Method} & En & Lowest (Ar) & \textbf{Average} \\
    mBERT$^{LoRA}$ & 79.32/65.99	& 46.17/29.67	&  55.55$\pm$0.60 / 37.54$\pm$0.63 \\
    mBERT$^{LoRA}$+GU & 78.63/65.03 & 47.70/30.21	& \gbc\underline{56.52}$\pm$0.78 / \underline{37.72}$\pm$0.79 \\
    mBERT$^{LoRA}$+FUN & 78.80/65.40	& 47.12/29.98	&\gbc\textbf{56.65}$\pm$0.79 / \textbf{38.02}$\pm$0.97  \\
\midrule

XLM-R$^{LoRA}$ & 80.04/67.27	& 46.19/28.92	& 59.40$\pm$0.61 / 40.36$\pm$0.38  \\
    
XLM-R$^{LoRA}$+GU & 80.27/66.68 & 52.35/34.36	& \gbc\textbf{63.11}$\pm$0.35 / \textbf{43.43}$\pm$0.13 \\
    
XLM-R$^{LoRA}$+FUN & 
    80.51/67.18 & 52.39/33.85 & \gbc\underline{62.62}$\pm$0.50 / \underline{43.21}$\pm$0.45 
    \\

    \bottomrule
    \end{tabular}
    }
    \caption{Zero-shot transfer results across MLQA and XQuAD, with the LoRA adapter. We bold the highest and underline the second-highest average value.}
    \label{tab:lora}
    \vspace{-2.5mm}
\end{table}

\rparagraph{Future Perspectives of FUN}
While GU remains effective, FUN offers the opportunity to potentially break away from the strict top-down unfreezing schedule and experiment with networks that have parallel structures (e.g., dual-network structures) in future work. Nonetheless, as the main finding in this work, FUN (i) provides evidence for a theory-based justification of heuristic unfreezing schedules from prior work, and (ii) it leads us to extend our understanding of learning dynamics during fine-tuning with scheduled unfreezing.

\section{Conclusion}

In this work, we first investigated whether scheduled unfreezing algorithms can help with generalization in the zero-shot cross-lingual transfer setting, and close the gap between parameter-efficient task-adapter training and full fine-tuning. Our experiments showed that scheduled unfreezing was indeed successful in closing this gap.  Next, we investigated the training dynamics of scheduled unfreezing using the trace of the Fisher Information Matrix ($\Tr(F)$). Our experiments revealed a link between scheduled unfreezing, $\Tr(F)$ and generalization performance. Finally, we proposed a general scheduled unfreezing algorithm ($\Tr(F)$-based scheduled unfreezing, FUN) that achieves performance comparable to existing heuristic variants, across multiple models and datasets. We hope to look into utilizing scheduled unfreezing to improve the cross-lingual generalization capabilities of large language models in the future.

\section{Limitations}

In this paper, we work with the trace of the Fisher Information Matrix as the metric for studying learning dynamics. While we believe our experiments and conclusions are widely applicable, there may be other complex factors and theoretical metrics (such as the eigenvalue spectrum of $F$ or other matrix norms, etc.) that we could potentially investigate. Furthermore, our use of $\Tr(F)$ is connected to prior work on the ``critical learning period'' during the early stages of training neural networks~\cite{dnncritical,achille2022critical}, which could help us gain deeper theoretical insights into parameter-efficient training methods. We also speculate GU may not degrade the performance of full parameter fine-tuning~\cite{raffel2022t5} if it is done early in the training (i.e., $kL \ll N$) rather than evenly unfrozen throughout the entire training process (i.e. $k = N/L$). However, such investigations are outside of the scope of this work, and we leave it for future work. 

\section*{Acknowledgments}
This work was funded by the German Federal Ministry of Education and Research (BMBF) under the promotional reference 13N15897 (MISRIK), and the LOEWE initiative (Hesse, Germany) within the emergenCITY center. The work was also supported in part by a personal Royal Society University Research Fellowship (no 221137; 2022-) awarded to Ivan Vuli\'{c}.

We thank Sebastian Ruder, Ji-Ung Lee, Nico Daheim, and Tim Baumg{\"a}rtner for their valuable feedback and suggestions on a draft of this paper. We thank Kexin Wang and Alan Ansell for discussions about training adapters.

\bibliography{anthology,custom}

\begin{thebibliography}{42}
\expandafter\ifx\csname natexlab\endcsname\relax\def\natexlab#1{#1}\fi

\bibitem[{Achille et~al.(2019)Achille, Rovere, and Soatto}]{dnncritical}
Alessandro Achille, Matteo Rovere, and Stefano Soatto. 2019.
\newblock \href {https://openreview.net/forum?id=BkeStsCcKQ} {Critical learning periods in deep networks}.
\newblock In \emph{7th International Conference on Learning Representations, {ICLR} 2019, New Orleans, LA, USA, May 6-9, 2019}.

\bibitem[{Amari(1998)}]{natural-grad}
Shun-ichi Amari. 1998.
\newblock \href {https://doi.org/10.1162/089976698300017746} {Natural gradient works efficiently in learning}.
\newblock \emph{Neural Computation}, 10(2):251--276.

\bibitem[{Ansell et~al.(2021)Ansell, Ponti, Pfeiffer, Ruder, Glava{\v{s}}, Vuli{\'c}, and Korhonen}]{ansell-etal-2021-mad-g}
Alan Ansell, Edoardo~Maria Ponti, Jonas Pfeiffer, Sebastian Ruder, Goran Glava{\v{s}}, Ivan Vuli{\'c}, and Anna Korhonen. 2021.
\newblock \href {https://doi.org/10.18653/v1/2021.findings-emnlp.410} {{MAD}-{G}: {M}ultilingual adapter generation for efficient cross-lingual transfer}.
\newblock In \emph{Findings of the Association for Computational Linguistics: EMNLP 2021}, pages 4762--4781, Punta Cana, Dominican Republic. Association for Computational Linguistics.

\bibitem[{Artetxe et~al.(2020)Artetxe, Ruder, and Yogatama}]{artetxe-etal-2020-cross-xquad}
Mikel Artetxe, Sebastian Ruder, and Dani Yogatama. 2020.
\newblock \href {https://doi.org/10.18653/v1/2020.acl-main.421} {On the cross-lingual transferability of monolingual representations}.
\newblock In \emph{Proceedings of the 58th Annual Meeting of the Association for Computational Linguistics}, pages 4623--4637, Online. Association for Computational Linguistics.

\bibitem[{Bapna and Firat(2019)}]{bapna-firat-2019-simple}
Ankur Bapna and Orhan Firat. 2019.
\newblock \href {https://doi.org/10.18653/v1/D19-1165} {Simple, scalable adaptation for neural machine translation}.
\newblock In \emph{Proceedings of the 2019 Conference on Empirical Methods in Natural Language Processing and the 9th International Joint Conference on Natural Language Processing (EMNLP-IJCNLP)}, pages 1538--1548, Hong Kong, China. Association for Computational Linguistics.

\bibitem[{Conneau et~al.(2020)Conneau, Khandelwal, Goyal, Chaudhary, Wenzek, Guzm{\'a}n, Grave, Ott, Zettlemoyer, and Stoyanov}]{conneau-etal-2020-unsupervised}
Alexis Conneau, Kartikay Khandelwal, Naman Goyal, Vishrav Chaudhary, Guillaume Wenzek, Francisco Guzm{\'a}n, Edouard Grave, Myle Ott, Luke Zettlemoyer, and Veselin Stoyanov. 2020.
\newblock \href {https://doi.org/10.18653/v1/2020.acl-main.747} {Unsupervised cross-lingual representation learning at scale}.
\newblock In \emph{Proceedings of the 58th Annual Meeting of the Association for Computational Linguistics}, pages 8440--8451, Online. Association for Computational Linguistics.

\bibitem[{Conneau et~al.(2018)Conneau, Rinott, Lample, Williams, Bowman, Schwenk, and Stoyanov}]{conneau-etal-2018-xnli}
Alexis Conneau, Ruty Rinott, Guillaume Lample, Adina Williams, Samuel Bowman, Holger Schwenk, and Veselin Stoyanov. 2018.
\newblock \href {https://doi.org/10.18653/v1/D18-1269} {{XNLI}: Evaluating cross-lingual sentence representations}.
\newblock In \emph{Proceedings of the 2018 Conference on Empirical Methods in Natural Language Processing}, pages 2475--2485, Brussels, Belgium. Association for Computational Linguistics.

\bibitem[{Devlin et~al.(2019)Devlin, Chang, Lee, and Toutanova}]{devlin-etal-2019-bert}
Jacob Devlin, Ming-Wei Chang, Kenton Lee, and Kristina Toutanova. 2019.
\newblock \href {https://doi.org/10.18653/v1/N19-1423} {{BERT}: Pre-training of deep bidirectional transformers for language understanding}.
\newblock In \emph{Proceedings of the 2019 Conference of the North {A}merican Chapter of the Association for Computational Linguistics: Human Language Technologies, Volume 1 (Long and Short Papers)}, pages 4171--4186, Minneapolis, Minnesota. Association for Computational Linguistics.

\bibitem[{Fisher(1925)}]{Fisher1925TheoryOS}
Rory~A. Fisher. 1925.
\newblock Theory of statistical estimation.
\newblock \emph{Mathematical Proceedings of the Cambridge Philosophical Society}, 22:700 -- 725.

\bibitem[{Golatkar et~al.(2019)Golatkar, Achille, and Soatto}]{achille2019time}
Aditya Golatkar, Alessandro Achille, and Stefano Soatto. 2019.
\newblock \href {https://proceedings.neurips.cc/paper/2019/hash/87784eca6b0dea1dff92478fb786b401-Abstract.html} {Time matters in regularizing deep networks: Weight decay and data augmentation affect early learning dynamics, matter little near convergence}.
\newblock In \emph{Advances in Neural Information Processing Systems 32: Annual Conference on Neural Information Processing Systems 2019, NeurIPS 2019, December 8-14, 2019, Vancouver, BC, Canada}, pages 10677--10687. Curran Associates, Inc.

\bibitem[{He et~al.(2021)He, Liu, Gao, and Chen}]{deberta}
Pengcheng He, Xiaodong Liu, Jianfeng Gao, and Weizhu Chen. 2021.
\newblock \href {https://openreview.net/forum?id=XPZIaotutsD} {{DeBERTa}: decoding-enhanced bert with disentangled attention}.
\newblock In \emph{9th International Conference on Learning Representations, {ICLR} 2021, Online, Austria, May 3-7, 2021}.

\bibitem[{Houlsby et~al.(2019)Houlsby, Giurgiu, Jastrzebski, Morrone, De~Laroussilhe, Gesmundo, Attariyan, and Gelly}]{pmlr-v97-houlsby19a}
Neil Houlsby, Andrei Giurgiu, Stanislaw Jastrzebski, Bruna Morrone, Quentin De~Laroussilhe, Andrea Gesmundo, Mona Attariyan, and Sylvain Gelly. 2019.
\newblock \href {https://proceedings.mlr.press/v97/houlsby19a.html} {Parameter-efficient transfer learning for {NLP}}.
\newblock In \emph{Proceedings of the 36th International Conference on Machine Learning}, volume~97 of \emph{Proceedings of Machine Learning Research}, pages 2790--2799. PMLR.

\bibitem[{Howard and Ruder(2018)}]{gu}
Jeremy Howard and Sebastian Ruder. 2018.
\newblock \href {https://doi.org/10.18653/v1/P18-1031} {Universal language model fine-tuning for text classification}.
\newblock In \emph{Proceedings of the 56th Annual Meeting of the Association for Computational Linguistics (Volume 1: Long Papers)}, pages 328--339, Melbourne, Australia. Association for Computational Linguistics.

\bibitem[{Hu et~al.(2022)Hu, Shen, Wallis, Allen{-}Zhu, Li, Wang, Wang, and Chen}]{hu2022lora}
Edward~J. Hu, Yelong Shen, Phillip Wallis, Zeyuan Allen{-}Zhu, Yuanzhi Li, Shean Wang, Lu~Wang, and Weizhu Chen. 2022.
\newblock \href {https://openreview.net/forum?id=nZeVKeeFYf9} {Lo{RA}: Low-rank adaptation of large language models}.
\newblock In \emph{The Tenth International Conference on Learning Representations, {ICLR} 2022, Virtual Event, April 25-29, 2022}. OpenReview.net.

\bibitem[{Hu et~al.(2020)Hu, Ruder, Siddhant, Neubig, Firat, and Johnson}]{xtreme}
Junjie Hu, Sebastian Ruder, Aditya Siddhant, Graham Neubig, Orhan Firat, and Melvin Johnson. 2020.
\newblock \href {http://proceedings.mlr.press/v119/hu20b.html} {{XTREME:} {A} massively multilingual multi-task benchmark for evaluating cross-lingual generalisation}.
\newblock In \emph{Proceedings of the 37th International Conference on Machine Learning, {ICML} 2020, 13-18 July 2020, Virtual Event}, pages 4411--4421.

\bibitem[{Jastrzebski et~al.(2021)Jastrzebski, Arpit, Astrand, Kerg, Wang, Xiong, Socher, Cho, and Geras}]{jastrzebski2021catastrophic}
Stanislaw Jastrzebski, Devansh Arpit, Oliver Astrand, Giancarlo~B Kerg, Huan Wang, Caiming Xiong, Richard Socher, Kyunghyun Cho, and Krzysztof~J Geras. 2021.
\newblock \href {https://proceedings.mlr.press/v139/jastrzebski21a.html} {Catastrophic fisher explosion: Early phase fisher matrix impacts generalization}.
\newblock In \emph{Proceedings of the 38th International Conference on Machine Learning}, volume 139 of \emph{Proceedings of Machine Learning Research}, pages 4772--4784. PMLR.

\bibitem[{Kirkpatrick et~al.(2017)Kirkpatrick, Pascanu, Rabinowitz, Veness, Desjardins, Rusu, Milan, Quan, Ramalho, Grabska-Barwinska, Hassabis, Clopath, Kumaran, and Hadsell}]{kirkpatrick17}
James Kirkpatrick, Razvan Pascanu, Neil Rabinowitz, Joel Veness, Guillaume Desjardins, Andrei~A. Rusu, Kieran Milan, John Quan, Tiago Ramalho, Agnieszka Grabska-Barwinska, Demis Hassabis, Claudia Clopath, Dharshan Kumaran, and Raia Hadsell. 2017.
\newblock \href {https://doi.org/10.1073/pnas.1611835114} {Overcoming catastrophic forgetting in neural networks}.
\newblock \emph{Proceedings of the National Academy of Sciences}, 114(13):3521--3526.

\bibitem[{Kleinman et~al.(2023)Kleinman, Achille, and Soatto}]{achille2022critical}
Michael Kleinman, Alessandro Achille, and Stefano Soatto. 2023.
\newblock \href {https://doi.org/10.1109/CVPR52729.2023.02327} {Critical learning periods for multisensory integration in deep networks}.
\newblock In \emph{{IEEE/CVF} Conference on Computer Vision and Pattern Recognition, {CVPR} 2023, Vancouver, BC, Canada, June 17-24, 2023}, pages 24296--24305. {IEEE}.

\bibitem[{Kumar et~al.(2022)Kumar, Raghunathan, Jones, Ma, and Liang}]{kumar2022finetuning}
Ananya Kumar, Aditi Raghunathan, Robbie~Matthew Jones, Tengyu Ma, and Percy Liang. 2022.
\newblock \href {https://openreview.net/forum?id=UYneFzXSJWh} {Fine-tuning can distort pretrained features and underperform out-of-distribution}.
\newblock In \emph{10th International Conference on Learning Representations, {ICLR} 2019, Online, Apr 25-29, 2022}.

\bibitem[{Kunstner et~al.(2019)Kunstner, Hennig, and Balles}]{kunstner2019empirical}
Frederik Kunstner, Philipp Hennig, and Lukas Balles. 2019.
\newblock \href {https://proceedings.neurips.cc/paper/2019/file/46a558d97954d0692411c861cf78ef79-Paper.pdf} {Limitations of the empirical fisher approximation for natural gradient descent}.
\newblock In \emph{Advances in Neural Information Processing Systems 32: Annual Conference on Neural Information Processing Systems 2019, NeurIPS 2019, December 8-14, 2019, Vancouver, BC, Canada}, pages 4158--4169. Curran Associates, Inc.

\bibitem[{Lee et~al.(2023)Lee, Chen, Tajwar, Kumar, Yao, Liang, and Finn}]{liang2022surgical}
Yoonho Lee, Annie~S. Chen, Fahim Tajwar, Ananya Kumar, Huaxiu Yao, Percy Liang, and Chelsea Finn. 2023.
\newblock \href {https://openreview.net/pdf?id=APuPRxjHvZ} {Surgical fine-tuning improves adaptation to distribution shifts}.
\newblock In \emph{The Eleventh International Conference on Learning Representations, {ICLR} 2023, Kigali, Rwanda, May 1-5, 2023}. OpenReview.net.

\bibitem[{Lewis et~al.(2020)Lewis, Oguz, Rinott, Riedel, and Schwenk}]{lewis-etal-2020-mlqa}
Patrick Lewis, Barlas Oguz, Ruty Rinott, Sebastian Riedel, and Holger Schwenk. 2020.
\newblock \href {https://doi.org/10.18653/v1/2020.acl-main.653} {{MLQA}: Evaluating cross-lingual extractive question answering}.
\newblock In \emph{Proceedings of the 58th Annual Meeting of the Association for Computational Linguistics}, pages 7315--7330, Online. Association for Computational Linguistics.

\bibitem[{Lodha et~al.(2023)Lodha, Belapurkar, Chalkapurkar, Tao, Ghosh, Basu, Petrov, and Srinivasan}]{lodha-etal-2023-surgical}
Abhilasha Lodha, Gayatri Belapurkar, Saloni Chalkapurkar, Yuanming Tao, Reshmi Ghosh, Samyadeep Basu, Dmitrii Petrov, and Soundararajan Srinivasan. 2023.
\newblock \href {https://doi.org/10.18653/v1/2023.findings-emnlp.204} {On surgical fine-tuning for language encoders}.
\newblock In \emph{Findings of the Association for Computational Linguistics: EMNLP 2023}, pages 3105--3113, Singapore. Association for Computational Linguistics.

\bibitem[{Loshchilov and Hutter(2019)}]{adamw}
Ilya Loshchilov and Frank Hutter. 2019.
\newblock \href {https://openreview.net/forum?id=Bkg6RiCqY7} {Decoupled weight decay regularization}.
\newblock In \emph{7th International Conference on Learning Representations, {ICLR} 2019, New Orleans, LA, USA, May 6-9, 2019}.

\bibitem[{Martens(2020)}]{naturalgrad}
James Martens. 2020.
\newblock \href {http://jmlr.org/papers/v21/17-678.html} {New insights and perspectives on the natural gradient method}.
\newblock \emph{Journal of Machine Learning Research}, 21(146):1--76.

\bibitem[{McCloskey and Cohen(1989)}]{MCCLOSKEY1989109}
Michael McCloskey and Neal~J. Cohen. 1989.
\newblock \href {https://doi.org/https://doi.org/10.1016/S0079-7421(08)60536-8} {Catastrophic interference in connectionist networks: The sequential learning problem}.
\newblock volume~24 of \emph{Psychology of Learning and Motivation}, pages 109--165. Academic Press.

\bibitem[{Parovi{\'c} et~al.(2022)Parovi{\'c}, Glava{\v{s}}, Vuli{\'c}, and Korhonen}]{parovic-etal-2022-bad}
Marinela Parovi{\'c}, Goran Glava{\v{s}}, Ivan Vuli{\'c}, and Anna Korhonen. 2022.
\newblock \href {https://doi.org/10.18653/v1/2022.naacl-main.130} {{BAD}-{X}: Bilingual adapters improve zero-shot cross-lingual transfer}.
\newblock In \emph{Proceedings of the 2022 Conference of the North American Chapter of the Association for Computational Linguistics: Human Language Technologies}, pages 1791--1799, Seattle, United States. Association for Computational Linguistics.

\bibitem[{Pfeiffer et~al.(2020)Pfeiffer, Vuli{\'c}, Gurevych, and Ruder}]{mad-x}
Jonas Pfeiffer, Ivan Vuli{\'c}, Iryna Gurevych, and Sebastian Ruder. 2020.
\newblock \href {https://doi.org/10.18653/v1/2020.emnlp-main.617} {{MAD-X}: {A}n {A}dapter-{B}ased {F}ramework for {M}ulti-{T}ask {C}ross-{L}ingual {T}ransfer}.
\newblock In \emph{Proceedings of the 2020 Conference on Empirical Methods in Natural Language Processing (EMNLP)}, pages 7654--7673, Online. Association for Computational Linguistics.

\bibitem[{Pfeiffer et~al.(2021)Pfeiffer, Vuli{\'c}, Gurevych, and Ruder}]{pfeiffer-etal-2021-unks}
Jonas Pfeiffer, Ivan Vuli{\'c}, Iryna Gurevych, and Sebastian Ruder. 2021.
\newblock \href {https://doi.org/10.18653/v1/2021.emnlp-main.800} {{UNK}s everywhere: {A}dapting multilingual language models to new scripts}.
\newblock In \emph{Proceedings of the 2021 Conference on Empirical Methods in Natural Language Processing}, pages 10186--10203, Online and Punta Cana, Dominican Republic. Association for Computational Linguistics.

\bibitem[{Ponti et~al.(2020)Ponti, Glava{\v{s}}, Majewska, Liu, Vuli{\'c}, and Korhonen}]{ponti-etal-2020-xcopa}
Edoardo~Maria Ponti, Goran Glava{\v{s}}, Olga Majewska, Qianchu Liu, Ivan Vuli{\'c}, and Anna Korhonen. 2020.
\newblock \href {https://doi.org/10.18653/v1/2020.emnlp-main.185} {{XCOPA}: A multilingual dataset for causal commonsense reasoning}.
\newblock In \emph{Proceedings of the 2020 Conference on Empirical Methods in Natural Language Processing (EMNLP)}, pages 2362--2376, Online. Association for Computational Linguistics.

\bibitem[{Raffel et~al.(2022)Raffel, Shazeer, Roberts, Lee, Narang, Matena, Zhou, Li, and Liu}]{raffel2022t5}
Colin Raffel, Noam Shazeer, Adam Roberts, Katherine Lee, Sharan Narang, Michael Matena, Yanqi Zhou, Wei Li, and Peter~J. Liu. 2022.
\newblock \href {https://dl.acm.org/doi/abs/10.5555/3455716.3455856} {Exploring the limits of transfer learning with a unified text-to-text transformer}.
\newblock \emph{Journal of Machine Learning Research}, 21(1).

\bibitem[{Rajpurkar et~al.(2016)Rajpurkar, Zhang, Lopyrev, and Liang}]{squad}
Pranav Rajpurkar, Jian Zhang, Konstantin Lopyrev, and Percy Liang. 2016.
\newblock \href {https://doi.org/10.18653/v1/D16-1264} {{SQ}u{AD}: 100,000+ questions for machine comprehension of text}.
\newblock In \emph{Proceedings of the 2016 Conference on Empirical Methods in Natural Language Processing}, pages 2383--2392, Austin, Texas. Association for Computational Linguistics.

\bibitem[{Ramponi and Plank(2020)}]{ramponi-plank-2020-neural}
Alan Ramponi and Barbara Plank. 2020.
\newblock \href {https://doi.org/10.18653/v1/2020.coling-main.603} {Neural unsupervised domain adaptation in {NLP}{---}{A} survey}.
\newblock In \emph{Proceedings of the 28th International Conference on Computational Linguistics}, pages 6838--6855, Barcelona, Spain (Online). International Committee on Computational Linguistics.

\bibitem[{Roemmele et~al.(2011)Roemmele, Bejan, and Gordon}]{copa}
Melissa Roemmele, Cosmin~A. Bejan, and Andrew~S. Gordon. 2011.
\newblock \href {https://people.ict.usc.edu/~gordon/publications/AAAI-SPRING11A.PDF} {Choice of plausible alternatives: An evaluation of commonsense causal reasoning}.
\newblock In \emph{AAAI Spring Symposium on Logical Formalizations of Commonsense Reasoning}, Stanford University.

\bibitem[{Shun-ichi and Hiroshi(2000)}]{amari2000infogeo}
Amari Shun-ichi and Nagaoka Hiroshi. 2000.
\newblock Methods of information geometry.
\newblock volume 191 of \emph{Translations of Mathematical Monographs}.

\bibitem[{Singh and Alistarh(2020)}]{fishprune}
Sidak~Pal Singh and Dan Alistarh. 2020.
\newblock \href {https://proceedings.neurips.cc/paper/2020/hash/d1ff1ec86b62cd5f3903ff19c3a326b2-Abstract.html} {Woodfisher: Efficient second-order approximation for neural network compression}.
\newblock In \emph{Advances in Neural Information Processing Systems 33: Annual Conference on Neural Information Processing Systems 2020, NeurIPS 2020, December 6-12, 2020, Online}. Curran Associates, Inc.

\bibitem[{Stickland and Murray(2019)}]{pmlr-v97-stickland19a}
Asa~Cooper Stickland and Iain Murray. 2019.
\newblock \href {https://proceedings.mlr.press/v97/stickland19a.html} {{BERT} and {PAL}s: Projected attention layers for efficient adaptation in multi-task learning}.
\newblock In \emph{Proceedings of the 36th International Conference on Machine Learning}, volume~97 of \emph{Proceedings of Machine Learning Research}, pages 5986--5995. PMLR.

\bibitem[{Sung et~al.(2021)Sung, Nair, and Raffel}]{sung2021training}
Yi{-}Lin Sung, Varun Nair, and Colin Raffel. 2021.
\newblock \href {https://proceedings.neurips.cc/paper/2021/hash/cb2653f548f8709598e8b5156738cc51-Abstract.html} {Training neural networks with fixed sparse masks}.
\newblock In \emph{Advances in Neural Information Processing Systems 34: Annual Conference on Neural Information Processing Systems 2021, NeurIPS 2021, December 6-14, 2021, Online}, pages 24193--24205. Curran Associates, Inc.

\bibitem[{Swayamdipta et~al.(2020)Swayamdipta, Schwartz, Lourie, Wang, Hajishirzi, Smith, and Choi}]{swayamdipta-etal-2020-dataset}
Swabha Swayamdipta, Roy Schwartz, Nicholas Lourie, Yizhong Wang, Hannaneh Hajishirzi, Noah~A. Smith, and Yejin Choi. 2020.
\newblock \href {https://doi.org/10.18653/v1/2020.emnlp-main.746} {Dataset cartography: Mapping and diagnosing datasets with training dynamics}.
\newblock In \emph{Proceedings of the 2020 Conference on Empirical Methods in Natural Language Processing (EMNLP)}, pages 9275--9293, Online. Association for Computational Linguistics.

\bibitem[{Williams et~al.(2018)Williams, Nangia, and Bowman}]{williams-etal-2018-broad}
Adina Williams, Nikita Nangia, and Samuel Bowman. 2018.
\newblock \href {https://doi.org/10.18653/v1/N18-1101} {A broad-coverage challenge corpus for sentence understanding through inference}.
\newblock In \emph{Proceedings of the 2018 Conference of the North {A}merican Chapter of the Association for Computational Linguistics: Human Language Technologies, Volume 1 (Long Papers)}, pages 1112--1122, New Orleans, Louisiana. Association for Computational Linguistics.

\bibitem[{Xu et~al.(2021)Xu, Luo, Zhang, Tan, Chang, Huang, and Huang}]{childtuning}
Runxin Xu, Fuli Luo, Zhiyuan Zhang, Chuanqi Tan, Baobao Chang, Songfang Huang, and Fei Huang. 2021.
\newblock \href {https://doi.org/10.18653/v1/2021.emnlp-main.749} {Raise a child in large language model: Towards effective and generalizable fine-tuning}.
\newblock In \emph{Proceedings of the 2021 Conference on Empirical Methods in Natural Language Processing}, pages 9514--9528, Punta Cana, Dominican Republic. Association for Computational Linguistics.

\bibitem[{Yang et~al.(2022)Yang, Ding, Guo, Lv, and Tang}]{parameffhead}
Zhuoyi Yang, Ming Ding, Yanhui Guo, Qingsong Lv, and Jie Tang. 2022.
\newblock \href {https://preview.aclanthology.org/emnlp-22-ingestion/2022.emnlp-main.514.pdf} {Parameter-efficient tuning makes a good classification head}.
\newblock In \emph{Proceedings of the 2022 Conference on Empirical Methods in Natural Language Processing}, Abu Dahbi, United Arab Emirates. Association for Computational Linguistics.

\end{thebibliography}

\clearpage 
\appendix

\section{Adapter Architecture}\label{app:adapter}

\begin{figure}
    \centering
    \includegraphics[width=0.75\linewidth]{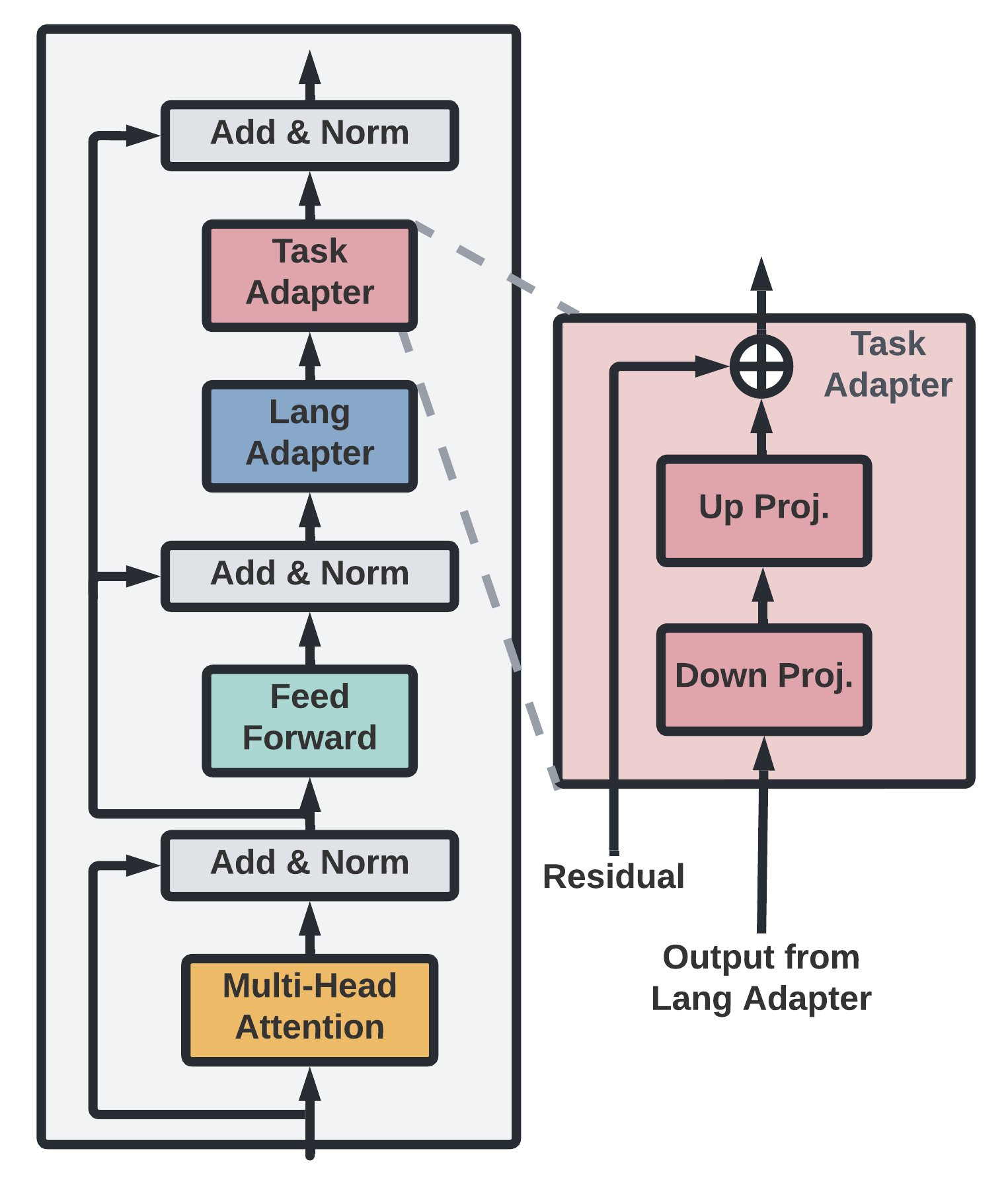}
    \caption{Adapter Architecture.}
    \label{fig:adapters}
\end{figure}

Figure~\ref{fig:adapters} shows the adapter architecture for cross-lingual transfer based on MAD-X. Each adapter consists of a down-projection followed by a ReLU activation and an up-projection, inserted after the feed-forward layer in every Transformer layer.

\section{Hyperparameters}\label{app:hparams}
We include the detailed hyperparameters used in our experiments in Table~\ref{tab:hparams} and~\ref{tab:hparams_fs}. We use the AdamW optimizer~\cite{adamw} for all experiments, without warmups and a maximum gradient norm of 1 in all models.

For COPA we train the model with longer epochs for scheduled unfreezing. since the training data for COPA only contains 400 instances, the training time is still very small (<1 hour). We found standard training for task adapters for COPA does not benefit from longer training time and a smaller learning rate.

In addition, some of the language adapters are missing from the AdapterHub (just for mBERT or both for mBERT and XLM-R). We follow the same adapter configuration from~\cite{mad-x}, and train those language adapters that are missing for mBERT (Italian, Tamil and Thai) with the Wikipedia data, a learning rate of 1e-4 , batch size of 64, sequence length of 512, and a maximum 100 epochs/training budget or 24 hours (whichever is reached first). 

The task adapters have a reduction factor of 16 as indicated in MAD-X.

\begin{table}[h]
\setlength\tabcolsep{2pt}
 \resizebox{0.99\linewidth}{!}{
    \centering
    \begin{tabular}{l|cc cccc}
    \toprule
     Model & Epochs & lr & batchsize & k-LPFT & k-GU & k-FUN \\
    \midrule
    \multicolumn{7}{c}{\textit{SQuAD} - LR Schedule: linear} \\
    \midrule
        mBERT & 5 &  5e-4 & 32 & 800 & 800 & 800\\
        XLM-R & 15 & 2e-4/5e-4 & 32 & 800 & 800 & 100\\
    \midrule
    \multicolumn{7}{c}{\textit{COPA}- LR Schedule: constant} \\
    \midrule
        mBERT & 500/5000 & 1e-4/1e-5 & 64 & 800 & 50  & 50 \\
        XLM-R & 500/5000 & 1e-4/1e-5 & 64 & 800 & 1000 & 1000 \\
    \midrule
    \multicolumn{7}{c}{\textit{MNLI}- LR Schedule: linear} \\
    \midrule
        mBERT & 15 & 5e-4 & 128 & 25 & 800 & 50\\
        XLM-R & 15 & 5e-4 & 128 & 800 & 800 & 800 \\
    \bottomrule
    \end{tabular}
        }
    \caption{Hyperparameters used in the main experiments. $x/y$ denotes: $x$ for standard adapter training and $y$ for all scheduled unfreezing experiments.}
    \label{tab:hparams}
\end{table}

\begin{table}[]
\setlength\tabcolsep{2pt}
 \resizebox{0.99\linewidth}{!}{
    \centering
    \begin{tabular}{l|cc cccc}
    \toprule
     Model & Epochs & lr & batchsize & k-1k & k-5k & k-10k  \\
    \midrule
    \multicolumn{7}{c}{\textit{SQuAD}} \\
    \midrule
        mBERT & 20 & 5e-4 & 32 & 10  & 50 & 50\\
        XLM-R & 20 & 5e-4 & 32 & 10  & 50 & 50\\
    \midrule
    \multicolumn{7}{c}{\textit{MNLI}} \\
    \midrule
        mBERT & 50 & 5e-4 & 128  & 1 & 25 & 25\\
        XLM-R & 50 & 5e-4 & 128  & 1 & 25 & 10 \\
    \bottomrule
    \end{tabular}
        }
    \caption{Hyperparameters used in the experiments with reduced task training data.}
    \label{tab:hparams_fs}

\end{table}

\section{Dataset Statistics}\label{app:data_stats}

We include the dataset statistics in Table~\ref{tab:data_stats}. The training data for XQuAD and MLQA are SQuAD~\cite{squad}. The training data for XCOPA is COPA~\cite{copa} and the training data for XNLI is MNLI~\cite{williams-etal-2018-broad}. All datasets used are available on HuggingFace. The language names and codes in our experiments are in Table~\ref{tab:lang_code}.

\begin{table}[h]
\setlength\tabcolsep{2pt}
 \resizebox{0.99\linewidth}{!}{
    \centering
    \begin{tabular}{l|cccr}
    \toprule
        Train data / Test data & n. train (En) & n. val (En) & n. test & n. lang  \\
    \midrule
        SQuAD / XQuAD & 87599 & 10570 & 1190 & 11 \\
        SQuAD / MLQA & 87599 & 10570 & 4517 -- 5495 & 5\\
        COPA / XCOPA & 400 & 100 & 500 & 11\\
        MNLI / XNLI & 392702 & 2490 & 5010 & 14\\
    \bottomrule
    \end{tabular}
        }
    \caption{Dataset statistics.}
    \label{tab:data_stats}

\end{table}

\begin{table}[h]
\setlength\tabcolsep{2pt}
 \resizebox{0.99\linewidth}{!}{
    \centering
    \begin{tabular}{lc|lc|lc}
    \toprule
        Language & Code & Language & Code & Language & Code \\
    \midrule
        Arabic & Ar & German & De  & Greek & El \\
        Spanish & Es & Hindi & Hi  & Russian & Ru \\ 
        Thai & Th & Turkish & Tr & Vietnamese & Vi \\ 
        Estonian & Et & Haitian & Ht  & Italian & It \\
        Indonesian & Id & Quechua & Qu & Swahili & Sw \\
        Chinese & Zh & Tamil & Ta \\
    \bottomrule
    \end{tabular}
        }
    \caption{Language code.}
    \label{tab:lang_code}

\end{table}

\section{Pseudo Code for $\Tr(F)$ calculation}\label{app:fim_calc}

We provide the pseudo code for $\Tr(F)$ calculation in Algorithm~\ref{alg:fim}. Alternatively, please see our code at \url{URL}.
Let $\textrm{FORWARD}(*)$ be the standard forward pass operation, $\textrm{SAMPLE}(*)$ be a function sampling labels from the label distribution of data, and $\textrm{AGGAVG}(*)$ the function that aggregates $\Tr(F)$ by task adapter blocks then taking the average over the number of trainable layers.

\begin{algorithm}[h]
\caption{$\Tr(F)$ Calculation}\label{alg:fim}
\small
\begin{algorithmic}[1]
\Require {Number of batches $N$ to sample from training data $D$ for computing $\Tr(F)$, $\mathcal{P}$ trainable parameters.}
\Statex
\State Copy the model. \Comment{Avoid interference to the standard optimization.}
\For{$i = 1 \dots \text{N}$}
    \State Sample a data batch $b \sim D$ 
    \State $\textrm{outputs} = \textrm{FORWARD}(*)$
    \State $\textrm{labels} = \textrm{SAMPLE}(*)$ 
    \State prob = $\textrm{LogSoftmax}(\textrm{outputs})$
    \State $\textrm{loss} = \textrm{NLL}(\textrm{prob}; \textrm{labels})$
    \State loss.backward()
    \For{$p_j = 1 \dots |\mathcal{P}|$}
        \State $\Tr(F)_j = p_j.grad^2$ / $|b|$
    \EndFor
    \State $\Tr(F) = \textrm{AGGAVG}(*)$
\EndFor
\end{algorithmic}
\end{algorithm}

\section{Additional Discussion on Combining Scheduled Unfreezing with a Regularizer}\label{app:discussion}

Prior work~\cite{jastrzebski2021catastrophic} proposed a FIM regularizer that relies on an artificially low learning rate (i.e., 10\% of the optimal learning rate for standard training). Adapter training requires a higher learning rate in general (e.g., learning rate values in Table~\ref{tab:hparams}), hence it is not straightforward to use an FIM regularizer directly. In addition, the generalization experiments in~\citet{jastrzebski2021catastrophic} are considered in-distribution (ID), whereas we focus on distribution shifts (i.e., cross-lingual and OOD).

Recent work such as LPFT~\cite{kumar2022finetuning} shows that ID and OOD performances do not correlate well, which we also observe in cross-lingual evaluation, especially when only English data is available for training and validation in zero-shot cross-lingual transfer (trends in English monolingual results can be inconsistent with trends in cross-lingual results). 

Hence, although it is possible to combine scheduled unfreezing with a regularization method, it is not obvious how to do so in the best way. A more extensive investigation is outside of the scope of this paper and we reserve this topic for future work.

\section{Detailed Results for Experiments}\label{app:full_results}

We show the detailed per-language experimental results in Tables~\ref{tab:main_xquad_ada} to~\ref{tab:main_xnli_ada}.

The baselines (full parameter fine-tuning) results for plotting Figure~\ref{fig:rel_gain} are in Table~\ref{tab:baselines}.
\begin{table}[]
\setlength\tabcolsep{2pt}
    \centering
    \scriptsize
    \begin{tabular}{l|cc }
    \toprule
     Model & MLQA (F1) &  XQuAD (F1) \\
    \midrule
        mBERT & 56.85 & 63.33 \\
        XLM-R & 62.59 & 71.98\\
    \midrule
        Model & XCOPA (Acc.) &  XNLI (Acc.) \\
    \midrule
        mBERT & 53.39 & 63.60\\
        XLM-R & 54.99 & 73.43 \\
    \bottomrule
    \end{tabular}
    \caption{Baselines (full fine-tuning) results for cross-lingual transfer.}
    \label{tab:baselines}

\end{table}

\begin{table*}[h]
\setlength\tabcolsep{2pt}
      \resizebox{0.99\textwidth}{!}{
    \centering
    \tiny
    \begin{tabular}{l |c| cccccccccc c}
    \toprule
        MLQA & En (F1 / EM) & Ar & De & El & Es & Hi & Ru & Th & Tr & Vi & Zh & \textbf{Avg. F1 / EM}\\
    \midrule
    mBERT$^{Ada}$ & 78.99/65.85	& 45.76	& 59.47	& - & 64.60	& 46.91	& - & - & -  & 57.01	& 58.64	& 55.40$\pm$0.94 / 37.07$\pm$0.72 \\
    mBERT$^{Ada}$+Rand & 79.22/65.94	& 46.65	& 57.92	& - & 67.91	& 46.03	& - & -  & - & 57.75	& 59.34	& 55.93$\pm$0.21 / 37.54$\pm$0.31 \\
    mBERT$^{Ada}$+GU & 78.04/64.20	& 47.96	& 57.95	& - & 68.64	& 51.75 & - & -  & - & 58.95	& 58.95	& 57.37$\pm$0.32 / 38.27$\pm$0.27\\
    mBERT$^{Ada}$+FUN & 78.82/65.29	& 48.20	& 58.96 & - & 67.19	& 51.51 & - & -  & -	& 59.47	& 58.64	& 57.33$\pm$0.51 / 38.29$\pm$0.63 \\
    \midrule
    XLM-R$^{Ada}$ & 79.52/65.99 & 51.74 & 59.64 & - & 68.33 & 61.77	&  - & - & - & 64.85 & 61.88 & 61.31$\pm$0.46 / 42.10$\pm$0.42 \\
    XLM-R$^{Ada}$+Rand & 80.32/67.01	& 50.33	& 61.72	&  - & 69.98	& 60.08	&  -&  -&  -& 63.81	& 62.22	& 61.36$\pm$1.69 / 41.59$\pm$1.96\\
    XLM-R$^{Ada}$+GU & 80.37/66.77	&  55.16	& 61.30 & - 	& 70.36	& 63.75	&  - & - & - & 66.45	& 63.79	& 63.47$\pm$0.12 / 43.55$\pm$0.11\\
    XLM-R$^{Ada}$+FUN & 80.92/66.70	& 53.17	& 62.38	& - & 70.04	& 63.77 & -& -& -	& 65.38	& 63.86	& 63.10$\pm$0.79 / 43.37$\pm$0.51\\
    \midrule
    XQuAD & En (F1 / EM) & Ar & De & El & Es & Hi & Ru & Th & Tr & Vi & Zh & \textbf{Avg. F1 / EM}\\
    \midrule
    mBERT$^{Ada}$ & 83.58/71.74	& 57.95	& 71.20	& 57.60	& 73.11	& 53.75	& 70.05	& 34.53	& 50.15	& 68.38	& 69.56	& 60.63$\pm$1.04 / 43.90$\pm$0.85 \\    
    mBERT$^{Ada}$+Rand & 83.86/72.31	& 59.09	& 71.58	& 62.06	& 74.84	& 52.61	& 70.00	&
    38.91	& 48.49	& 69.29	& 69.92	& 61.68$\pm$0.33 / 47.42$\pm$0.55  \\
    mBERT$^{Ada}$+GU & 83.21/71.55	& 62.90	& 72.39	& 62.38	& 74.43	& 56.28	& 69.46	& 44.08	& 53.39	& 70.10	& 69.37	& 63.48$\pm$0.22 / 46.76$\pm$0.44 \\
    mBERT$^{Ada}$+FUN  & 83.71/71.83	& 62.24	& 72.74	& 62.19	& 74.05	& 56.92	& 69.74	& 42.55	& 53.40	& 69.56	& 69.13	& 63.25$\pm$0.26 / 49.09$\pm$0.48\\

    \midrule
    XLM-R$^{Ada}$ & 83.48/72.69	& 65.47	& 72.74 & 71.92	& 74.88	& 67.70	& 73.58	&
    66.53	& 66.36	& 72.38	& 69.36	& 70.09$\pm$0.60 / 53.77$\pm$0.40\\
    XLM-R$^{Ada}$+Rand & 84.76/73.74	& 63.69	& 74.40	& 71.25	& 76.28	& 65.09	& 73.77	& 64.93	& 64.85	& 72.06	& 73.54	& 69.99$\pm$1.47 / 52.06$\pm$2.04\\
    XLM-R$^{Ada}$+GU & 84.49/73.57	& 67.83	& 75.55	& 74.26	&
    77.42	& 70.46	& 75.52	& 69.52	& 68.53	& 75.88	& 75.39	& 73.04$\pm$0.22 / 55.93$\pm$0.15 \\
    XLM-R$^{Ada}$+FUN & 84.91/73.80 & 66.69	& 75.94	& 74.07	& 76.58	& 69.59	& 75.48	& 67.59	& 68.19	& 74.52	& 74.77	& 72.34$\pm$0.40 / 55.21$\pm$0.63\\

    \bottomrule
    \end{tabular}
    }
    \vspace{-0.5mm}
    \caption{Zero-shot transfer results (F1) for MLQA and XQuAD. Average is the cross-lingual average without English.}
    \label{tab:main_xquad_ada}
    \vspace{-2mm}
\end{table*}

\begin{table*}[h]
\setlength\tabcolsep{2pt}
      \resizebox{0.99\textwidth}{!}{
    \centering
    \tiny
    \begin{tabular}{l |c|cccccccccccc c}
    \toprule
    XCOPA & En & Et & Ht & It & Id & Qu & Sw & Zh & Ta & Th & Tr & Vi &  \textbf{Avg. Acc.}\\
    \midrule
    mBERT$^{Ada}$ & 63.80	& 54.20	& 53.04	& 50.16	& 53.84	& 53.12	&
54.16	& 59.08	& 52.56	& 51.68	& 54.52	& 57.56	& 53.99$\pm$0.49\\
    mBERT$^{Ada}$+Rand & 65.00	& 53.36	& 52.32	& 50.96	& 53.60	& 54.00	& 53.44	&
    58.64	& 50.92	& 51.96	& 55.04	& 57.96	& 53.84$\pm$0.71 \\
    mBERT$^{Ada}$+GU & 66.60	& 54.44	& 52.60	& 50.00	& 54.88	& 53.52	& 53.52	& 59.76	& 52.12	& 52.36	& 55.44	& 58.60	& 54.29$\pm$0.60 \\
    mBERT$^{Ada}$+FUN & 66.40	& 53.44	& 52.92	& 50.68	& 54.76	& 53.48	&
    54.40	& 59.00	& 51.36	& 51.08	& 54.32	& 58.36	& 53.98$\pm$0.64\\
    \midrule
    XLM-R$^{Ada}$ & 65.20	& 56.16	& 51.28 & 56.72	& 58.00	& 51.80	& 55.60	& 59.12	& 56.44	& 57.72	& 56.48	& 55.96	& 55.93$\pm$1.58 \\
    XLM-R$^{Ada}$+Rand & 67.20	& 57.08	& 52.12	& 57.80	& 60.72	& 53.32	& 56.24	&
    60.08	& 57.36	& 58.20	& 56.76	& 57.88	& 57.05$\pm$0.42 \\
    XLM-R$^{Ada}$+GU &  66.00	& 58.56	& 52.52 & 58.24	&
    62.04	& 53.96	& 56.88	& 61.36	& 59.00	& 60.08	& 58.52	& 59.52	& 58.24$\pm$1.11 \\
    XLM-R$^{Ada}$+FUN & 67.80 & 58.16	& 52.08	& 57.44	& 61.28	& 55.04	&
    56.36	& 61.64	& 57.84	& 61.04	& 58.80	& 59.48	& 58.11$\pm$0.94\\

    \bottomrule
    \end{tabular}
    }
    \vspace{-0.5mm}
    \caption{Zero-shot transfer results (Accuracy) for XCOPA. Average is the cross-lingual average without English.}
    \label{tab:main_xcopa_ada}
    \vspace{-2mm}
\end{table*}

\begin{table*}[h]
\setlength\tabcolsep{2pt}
      \resizebox{0.99\textwidth}{!}{
    \centering
    \tiny
    \begin{tabular}{l |c|cccccccccccccc c}
    \toprule
    XNLI & En & Ar &  De & El & Es &  Hi & Ru & Sw & Th & Tr  & Vi & Zh &  \textbf{Avg. Acc.}\\
    \midrule
    mBERT$^{Ada}$ & 82.05	& 42.09	 & 65.81	& 62.16	& 70.84	&
57.92	& 63.76	& 37.45	& 40.89	& 61.53	&  68.01	& 65.08	& 57.78$\pm$1.68 \\
    mBERT$^{Ada}$+Rand & 81.64	& 53.98	&  66.32	& 62.85	& 70.33		& 58.15	& 65.08	& 45.95	& 41.80	& 61.25		& 67.73	& 65.12	& 59.87$\pm$0.96\\
    mBERT$^{Ada}$+GU & 81.79	& 62.78 & 66.25	& 63.51	& 70.28	& 
58.89	& 65.74	& 54.06	 & 38.97 & 62.25 &  68.52	& 67.17	& 61.67$\pm$1.04 \\
 mBERT$^{Ada}$+FUN & 81.70	& 58.48	&   66.32	& 63.87	& 70.98		&
59.08	& 65.45	& 53.73	& 41.13	&61.89	&  68.25	&65.75	& 61.36$\pm$0.51 \\

    \midrule
    XLM-R$^{Ada}$ & 84.31	& 70.42	& 76.16	& 75.80	& 78.85	&  70.16	& 75.14	& 68.16	& 71.14	& 72.33	& 
    75.06	& 73.24	& 73.31$\pm$0.44\\
     XLM-R$^{Ada}$+Rand & 84.52	& 69.91	&  75.91	& 75.05	& 78.04	&  69.56	& 74.51	&
    67.29	& 70.40	& 72.05	&  74.25	& 72.48	& 72.68$\pm$0.56 \\
    XLM-R$^{Ada}$+GU & 84.24 & 70.22	& 75.92	& 75.7	& 78.32		& 70.61	& 75.70	& 68.24	& 71.99	& 71.97	&  75.36	& 73.82	& 73.44$\pm$0.24\\

     XLM-R$^{Ada}$+FUN &  84.72	& 70.58	&  76.17	& 75.68	& 78.29	&  69.75	& 75.42	& 67.48 &
    71.44	& 71.71	&  74.75	& 73.20	& 73.13$\pm$0.53 \\

    \bottomrule
    \end{tabular}
    }
    \vspace{-0.5mm}
    \caption{Zero-shot transfer results (Accuracy) for XNLI. Average is the cross-lingual average without English.}
    \label{tab:main_xnli_ada}
    \vspace{-2mm}
\end{table*}

\section{Additional Experiments}\label{app:more_exp}

\subsection{Simulated Low-data Scenario}
In order to simulate setups with fewer annotated data for training and analyze the impact of scheduled unfreezing in those setups, we sample 1k, 5k, and 10k training examples from SQuAD and MNLI. We evaluate GU against the standard adapter training baseline (Table~\ref{tab:lessshot}). With a smaller amount of training data, we still observe the advantages of GU over standard task adapter fine-tuning.

\begin{table}[]
    \centering
        \resizebox{0.89\linewidth}{!}{
    \scriptsize
    \begin{tabular}{l|ccc}
    \toprule
         XQuAD (F1) &  1K & 5K & 10K\\
         \midrule
         mBERT$^{Ada}$ & 45.27$\pm$0.59 & 52.58$\pm$0.81&55.89$\pm$1.08\\
         mBERT$^{Ada}$+GU&\textbf{45.93}$\pm$0.50  & \textbf{53.10}$\pm$0.35&\textbf{56.47}$\pm$0.74 \\
         XLM-R$^{Ada}$& 44.11$\pm$1.43 & 57.20$\pm$0.36 &61.75$\pm$0.68\\
         XLM-R$^{Ada}$+GU& \textbf{48.42}$\pm$1.20 &  \textbf{59.88}$\pm$1.51 & \textbf{65.28}$\pm$0.76\\
         \midrule
        XNLI (Accuracy) &  1K & 5K & 10K\\
        \midrule
         mBERT$^{Ada}$& 43.86$\pm$1.43 &49.68$\pm$0.73& 52.34$\pm$0.40\\
         mBERT$^{Ada}$+GU&\textbf{44.69}$\pm$0.61  &\textbf{51.67}$\pm$0.43& \textbf{53.95}$\pm$1.47\\
         
         XLM-R$^{Ada}$&  52.75$\pm$2.03 &\textbf{64.22}$\pm$1.04& 65.80$\pm$0.61 \\ 
         XLM-R$^{Ada}$+GU& \textbf{52.86}$\pm$1.38 &64.15$\pm$0.35 & \textbf{65.91}$\pm$0.56\\
    \bottomrule
    \end{tabular}
    }
    \caption{Cross-lingual transfer performance with sub-sampled English task data for task fine-tuning.}
    \label{tab:lessshot}
    \vspace{-3.mm}
\end{table}

\subsection{Reverse Gradual Unfreezing}
We briefly experimented (2 runs) with the reverse order (bottom-up) of gradual unfreezing on MLQA, XQuAD and XNLI. We include the results for bottom-up GU ($rev$) in Figure~\ref{tab:gu_reverse} and included the numbers from standard GU for reference. The cross-lingual transfer results are significantly lower than the standard GU; however, the English results are similar.

\begin{table}[h]
\setlength\tabcolsep{2pt}
    \resizebox{0.99\linewidth}{!}{
    \centering
    \tiny
    \begin{tabular}{l| c c}
    \toprule
    MLQA & En (F1) & \textbf{Avg. F1}  \\
    \midrule
    mBERT$^{Ada}$+GU & 78.04 & 57.37 \\
    mBERT$^{Ada}$+GU (rev) & 78.71 & 49.09\\
    \midrule
    XLM-R$^{Ada}$+GU & 80.37	& 63.47 \\
    XLM-R$^{Ada}$+GU (rev) & 81.38	& 57.59 \\
    \midrule
    XQuAD & En (F1) & \textbf{Avg. F1}  \\
    \midrule
    mBERT$^{Ada}$+GU  & 83.21	& 63.48\\
    mBERT$^{Ada}$+GU (rev)  & 82.17 & 53.43 \\
    \midrule
    XLM-R$^{Ada}$+GU &  84.49	& 73.04 \\
        XLM-R$^{Ada}$+GU (rev) & 84.12	& 65.44 \\
   \midrule
   XNLI &   En (Acc.) &  \textbf{Avg. Acc.}
        \\
    \midrule
    mBERT$^{Ada}$+GU &  81.79	& 61.67\\
    mBERT$^{Ada}$+GU &  81.43	& 55.67 \\

    \midrule
    XLM-R$^{Ada}$+GU &  84.24 & 73.44 \\
    XLM-R$^{Ada}$+GU (rev)  & 84.23 & 72.50 \\
    \bottomrule
    \end{tabular}
    }
    \vspace{-0.5mm}
    \caption{Zero-shot transfer results of gradual unfreezing in reverse order across three datasets: MLQA, XQuAD, and XNLI. Average is the cross-lingual average without English.}
    \label{tab:gu_reverse}
    \vspace{-2mm}
\end{table}

\subsection{Experiments on Smaller Task Training Data with $\Tr(F)$-based Scheduling}

We additionally performed the same experiments with smaller training data as described in our main paper, but now with $\Tr(F)$-based scheduling ($+FUN$). The results are in Table~\ref{tab:lessshot_fim}. Our results show that FUN is also comparable to GU when there are fewer training instances available. The $k$ used in our experiment for FUN are (for 1k/5k/10k correspondingly): mBERT-XQuAD = [10,50,50], XLM-R-XQuAD = [10,50,25], mBERT-XNLI = [10,50,25], XLM-R-XNLI = [1,25,10]. The remaining hyperparameters are the same as in all other experiments.

\begin{table}[]
    \centering
    \scriptsize
    \begin{tabular}{l|ccc}
    \toprule
         XQuAD ((F1)) &  1K & 5K & 10K\\
         \midrule
         mBERT$^{Ada}$ & 45.27$\pm$0.59 & 52.58$\pm$0.81&55.89$\pm$1.08\\
         mBERT$^{Ada}$+GU&45.93$\pm$0.50  & 53.10$\pm$0.35&56.47$\pm$0.74 \\
         mBERT$^{Ada}$+FUN& 46.30$\pm$0.71& 53.68$\pm$0.38 & 56.50$\pm$0.97\\
        \midrule

         XLM-R$^{Ada}$& 44.11$\pm$1.43 & 57.20$\pm$0.36 &61.75$\pm$0.68\\
         XLM-R$^{Ada}$+GU& 48.42$\pm$1.20 &  59.88$\pm$1.51 & 65.28$\pm$0.76\\
         XLM-R$^{Ada}$+FUN& 47.67$\pm$1.73& 60.72$\pm$1.07&65.16$\pm$1.16  \\
         \midrule
        XNLI (Accuracy) &  1K & 5K & 10K\\
        \midrule
         mBERT$^{Ada}$& 43.86$\pm$1.43 &49.68$\pm$0.73& 52.34$\pm$0.40\\
         mBERT$^{Ada}$+GU&44.69$\pm$0.61  &51.67$\pm$0.43& 53.95$\pm$1.47\\
         mBERT$^{Ada}$+FUN& 44.86$\pm$0.49 &51.61$\pm$0.58 & 53.40$\pm$0.93\\
        \midrule
         XLM-R$^{Ada}$&  52.75$\pm$2.03 &64.22$\pm$1.04& 65.80$\pm$0.61 \\ 
         XLM-R$^{Ada}$+GU& 52.86$\pm$1.38 &64.15$\pm$0.35 & 65.91$\pm$0.56\\
        XLM-R$^{Ada}$+FUN& 52.29$\pm$1.85 &64.10$\pm$0.92 & 66.26$\pm$0.87\\
        
    \bottomrule
    \end{tabular}
    \caption{Cross-lingual transfer performance with sub-
sampled English task data for task fine-tuning.}
    \label{tab:lessshot_fim}
\end{table}

\clearpage
\subsection{Additional Results with LoRA Adapters}\label{app:lora}

We provide additional results with LoRA adapters in Table~\ref{tab:app_lora}. For simplicity and to draw comparisons to our experiments with MAD-X, we couple the unfreezing for query and value LoRAs (applied to `q' and `v' attentions) and use the default LoRA configuration [lora\_r=8, lora\_alpha=8] from AdapterHub. 

From the results, we can see that GU and FUN are consistently comparable. GU is slightly worse than the standard training likely due to the extremely small training data size of COPA. Overall, scheduled unfreezing algorithms can be easily applied to different adapter architectures to provide a performance boost.

The hyperparameters used in this experiment are in Table~\ref{tab:lora_hype}, where we kept the number of epochs for training and batch size the same as in our main experiments (Table~\ref{tab:hparams}). 

\begin{table*}[]
\setlength\tabcolsep{2pt}
    \resizebox{0.99\linewidth}{!}{
    \centering
    \tiny
    \begin{tabular}{l| ccc | ccc }
    \toprule
    \multicolumn{1}{c}{} & \multicolumn{3}{c}{\textit{MLQA (F1 / EM)}} & \multicolumn{3}{c}{\textit{XQuAD (F1 / EM)}} \\
    \midrule
    \textbf{Method} & En & Lowest (Th) & \textbf{Average} & En & Lowest (Ar) & \textbf{Average} \\

    mBERT$^{LoRA}$ & 79.32/65.99 & 46.17/29.67	&  55.55$\pm$0.60 / 37.54$\pm$0.63 
    & 83.68/71.50	& 37.56/29.44	&  61.53$\pm$0.37 / 45.01$\pm$0.44 \\
    mBERT$^{LoRA}$+GU & 78.63/65.03 & 47.70/30.21	& \gbc\underline{56.52}$\pm$0.78 / \underline{37.72}$\pm$0.79 
    & 84.24/72.75 & 40.73/31.41	& \gbc\textbf{63.12}$\pm$0.20 / \textbf{45.96}$\pm$0.38 \\
    mBERT$^{LoRA}$+FUN & 78.80/65.40 & 47.12/29.98	&\gbc\textbf{56.65}$\pm$0.79 / \textbf{38.02}$\pm$0.97  
    & 83.58/71.83	& 39.88/28.59	&\gbc\underline{62.92}$\pm$0.46 / \underline{45.80}$\pm$0.42 \\
    
    \midrule
    \textbf{Method} & En & Lowest (Ar) & \textbf{Average} & En & Lowest (Ar) & \textbf{Average} \\
    XLM-R$^{LoRA}$ & 80.04/67.27	& 46.19/28.92	& 59.40$\pm$0.61 / 40.36$\pm$0.38  & 83.35/72.27	& 58.09/42.10 	& 68.92$\pm$1.16 / 51.50$\pm$1.53  \\
    
    XLM-R$^{LoRA}$+GU & 80.27/66.68 & 52.35/34.36	& \gbc\textbf{63.11}$\pm$0.35 / \textbf{43.43}$\pm$0.13 
    & 84.70/73.12 & 66.72/49.50 & \gbc\textbf{72.27}$\pm$0.12 / \textbf{54.67}$\pm$0.34  \\
    
    XLM-R$^{LoRA}$+FUN & 
    80.51/67.18 & 52.39/33.85 & \gbc\underline{62.62}$\pm$0.50 / \underline{43.21}$\pm$0.45 
    & 84.22/72.29 & 65.69/48.62 & \gbc\underline{72.13}$\pm$0.18 / \underline{54.53}$\pm$0.14 
    \\

   \midrule   
   \multicolumn{1}{c}{} & \multicolumn{3}{c}{\textit{XCOPA (Accuracy)}} & \multicolumn{3}{c}{\textit{XNLI (Accuracy)}} \\
    \midrule
    \textbf{Method} & En & Lowest  & \textbf{Average} & En & Lowest (Sw)  & \textbf{Average} \\
    mBERT$^{LoRA}$ & 68.00	& 50.76	(Et) &  53.43$\pm$1.05 
    & 81.86	& 49.96 &  65.27$\pm$0.02 \\
    mBERT$^{LoRA}$+GU & 66.00 & 52.04 (Ta)	& \gbc \underline{54.67}$\pm$0.45 
    & 81.27 & 49.93 & \gbc \textbf{65.33}$\pm$0.15  \\
    mBERT$^{LoRA}$+FUN & 66.80	& 51.84 (Sw) & \gbc \textbf{54.83}$\pm$0.26 & 81.05 & 50.23 & \gbc \underline{65.32}$\pm$0.15 \\
    
    \midrule 
    \textbf{Method} & En & Lowest & \textbf{Average} & En & Lowest (Sw) & \textbf{Average} \\
    XLM-R$^{LoRA}$ & 63.60	& 50.35 (Ht) &  \underline{55.46}$\pm$0.88 
    & 84.26 & 64.79 & 72.98$\pm$0.33  \\
    XLM-R$^{LoRA}$+GU & 64.00 & 51.10 (Ht) & \gbc 55.07$\pm$1.37  
    & 84.04  & 65.29  & \gbc \underline{73.43}$\pm$0.20 \\
    XLM-R$^{LoRA}$+FUN & 64.40	& 49.04 (Qu) & \gbc \textbf{55.62}$\pm$0.99
    & 84.10  & 65.88  & \gbc \textbf{73.65}$\pm$0.57 \\

    \bottomrule
    \end{tabular}
    }
    \vspace{-0.5mm}
    \caption{Zero-shot transfer results with LoRA adapters. Average is the cross-lingual average without English. We bold the highest and underline the second-highest average value. \textit{Lowest} denotes the task performance for the lowest-performing target language per each evaluation dataset and base model.}
    \label{tab:app_lora}
    \vspace{-2mm}
\end{table*}

\begin{table}[h]
\setlength\tabcolsep{2pt}
 \resizebox{0.99\linewidth}{!}{
    \centering
    \begin{tabular}{l|cc ccc}
    \toprule
     Model & Epochs & lr & batchsize & k-GU & k-FUN \\
    \midrule
    \multicolumn{6}{c}{\textit{SQuAD} - LR Schedule: linear} \\
    \midrule
        mBERT & 5 &  5e-4/8e-4 & 32 & 800 & 800\\
        XLM-R & 15 & 5e-4 & 32 & 800 & 800\\
    \midrule
    \multicolumn{6}{c}{\textit{COPA - LR Schedule: constant}} \\
    \midrule
        mBERT & 500/5000 & 1e-4 & 64 & 50  & 100 \\
        XLM-R & 500/5000 & 1e-4 & 64 & 100 & 200 \\
    \midrule
    \multicolumn{6}{c}{\textit{MNLI} - LR Schedule: linear} \\
    \midrule
        mBERT & 15 & 5e-4 & 128 & 50 & 50\\
        XLM-R & 15 & 5e-4 & 128 & 800 & 800 \\
    \bottomrule
    \end{tabular}
        }
    \caption{Hyperparameters used in the experiments with LoRA adapters. $x/y$ denotes: $x$ for standard adapter training and $y$ for all scheduled unfreezing experiments.}
    \label{tab:lora_hype}
\end{table}

\subsection{Preliminary Results with mDeBERTa}
We include additional experiments on another, more recent base model, mDeBERTa~\cite{deberta}. mDeBERTa is the multilingual version of the recently proposed DeBERTa~\cite{deberta} model with disentangled attention to its word content and position representations. We used the (mdeberta-v3-base) model for all our experiments.

Note that we trained language adapters using MLM loss for the mDeBERTa model according to the setup described in~\cite{mad-x}. However, we see very large discrepancies in terms of transfer results for both XCOPA and XNLI when compared to the standard fine-tuning (the gaps are also much larger than the gaps for mBERT or XLM-R). We hypothesize that the discrepancies may be because mDeBERTa uses different attentions and the adapters we studied here are not designed for mDeBERTa (both in their architecture and in their training method). However, as tuning adapter architectures is beyond the scope of our study, we include the results here for completeness only.

\begin{table*}[h]
\setlength\tabcolsep{2pt}
      \resizebox{0.99\textwidth}{!}{
    \centering
    \tiny
    \begin{tabular}{l |c| cccccccccc c}
    \toprule
    MLQA & En (F1 / EM) & Ar & De & El & Es & Hi & Ru & Th & Tr & Vi & Zh & \textbf{Avg. F1 / EM}\\
    \midrule
    mDeBERTa$^{Ada}$ & 82.31 & 58.50 & 67.56 & -& 73.00	& 64.27	 & -& -& - & 64.92	& 65.27	& 65.59$\pm$0.22 / 46.19$\pm$0.29	\\
    mDeBERTa$^{Ada}$+GU & 82.27 & 61.19	& 67.01	& - & 73.26	& 65.71	&  - &  - &  - & 65.71	& 67.44	& 67.08$\pm$0.38 / 47.22$\pm$0.32  \\
    \midrule
    XQuAD & En (F1 / EM) & Ar & De & El & Es & Hi & Ru & Th & Tr & Vi & Zh & \textbf{Avg. F1 / EM}\\
    \midrule
    mDeBERTa$^{Ada}$ & 86.30	& 72.33	& 80.38	& 76.89	&
    79.93	& 72.77	& 77.27	& 69.97 & 71.51	& 74.99	&
    78.98	& 75.50$\pm$0.29 / 58.66$\pm$0.19\\
    mDeBERTa$^{Ada}$+GU & 85.52	& 73.31	& 79.59	& 78.41	& 79.77	& 73.58	& 77.87	& 70.98	& 72.66	& 75.42	& 79.15	& 76.07$\pm$0.13 / 58.90$\pm$0.12\\
    \bottomrule
    \end{tabular}
    }
    \vspace{-0.5mm}
    \caption{mDeBERTa: Zero-shot transfer results (F1) XQUAD and MLQA. Average is the cross-lingual average without English.}
    \label{tab:mdeberta_xquad}
    \vspace{-2mm}
\end{table*}

\begin{table*}[h]
\setlength\tabcolsep{2pt}
      \resizebox{0.99\textwidth}{!}{
    \centering
    \tiny
    \begin{tabular}{l |c|cccccccccccc c}
    \toprule
    XCOPA & En & Et & Ht & It & Id & Qu & Sw & Zh & Ta & Th & Tr & Vi &  \textbf{Avg. Acc.}\\
    \midrule
    mDeBERTa$^{Ada}$ & 65.75 & 55.32	& 55.68	& 58.64	& 61.00	& 51.40 & 56.48	& 62.76	& 57.16	& 57.28	& 55.76	& 59.04	& 57.32$\pm$4.46	\\
    mDeBERTa$^{Ada}$+GU &  65.80	& 57.96	& 56.60	& 59.36	&
    60.12	& 52.36	& 55.96	& 63.28	& 57.88	& 57.48	& 58.12	& 58.76	& 57.99$\pm$3.57\\
    \bottomrule
    \end{tabular}
    }
    \vspace{-0.5mm}
    \caption{mDeBERTa: Zero-shot transfer results (Accuracy) XCOPA. Average is the cross-lingual average without English.}
    \label{tab:mdeberta_xcopa}
    \vspace{-2mm}
\end{table*}

\begin{table*}[h]
\setlength\tabcolsep{2pt}
      \resizebox{0.99\textwidth}{!}{
    \centering
    \tiny
    \begin{tabular}{l |c|cccccccccccccc c}
    \toprule
    XNLI & En & Ar &  De & El & Es &Hi & Ru & Sw & Th & Tr & Vi & Zh &  \textbf{Avg. Acc.}\\
    \midrule
    mDeBERTa$^{Ada}$&     86.94	& 72.09	& 78.57	& 76.03	& 80.92		& 69.41	& 76.98	& 68.45	& 68.73	& 75.93	& 74.62	& 73.30	& 74.09$\pm$0.77 \\
    mDeBERTa$^{Ada}$+GU &  86.48	& 71.46		&
    77.90	& 75.50	& 79.23	&  67.30	& 75.84	&
    68.64	& 69.68	& 74.51	&  74.05	& 74.85	& 73.54$\pm$0.58\\
    \bottomrule
    \end{tabular}
    }
    \vspace{-0.5mm}
    \caption{mDeBERTa: Zero-shot transfer results (Accuracy) XNLI. Average is the cross-lingual average without English.}
    \label{tab:mdeberta_xnli}
    \vspace{-2mm}
\end{table*}

\end{document}